\documentclass{article} 
\usepackage{iclr2027_conference,times}

\usepackage{amsmath,amsfonts,bm}

\def\eqref#1{equation~\ref{#1}}

\def\1{\bm{1}}

\DeclareMathAlphabet{\mathsfit}{\encodingdefault}{\sfdefault}{m}{sl}
\SetMathAlphabet{\mathsfit}{bold}{\encodingdefault}{\sfdefault}{bx}{n}

\usepackage{hyperref}
\usepackage{url}
\usepackage{booktabs}
\usepackage{multirow}
\usepackage{graphicx}
\usepackage{caption}
\usepackage{wrapfig}
\usepackage{needspace}
\usepackage{caption}
\usepackage{float}
\usepackage{subcaption}

\title{EgoMatrix: Physically Faithful Dexterous Robot Demonstrations from Egocentric Human Videos}

\author{%
{\bfseries Ziheng Wang$^{1}$, Yanming Shao$^{2}$, Ting Mao$^{3}$}\\[0.2em]
{\bfseries Kaihan Chen$^{3}$, Yiqun Wang$^{4}$, Xuanye Wu$^{4}$, Yao Mu$^{1,4}$\thanks{Corresponding author}}\\[0.5em]
{\normalfont $^{1}$Shanghai Jiao Tong University \quad $^{2}$The University of Hong Kong}\\
{\normalfont $^{3}$Zhejiang University \quad $^{4}$Shanghai AI Laboratory}
}

\newcommand{\best}[1]{\textbf{#1}}
\newcommand{\second}[1]{\underline{#1}}

\iclrfinalcopy 
\begin{document}

\setcounter{footnote}{1}
\maketitle
\fancyhead{}
\renewcommand{\headrulewidth}{0pt}

\begin{abstract}
Egocentric human videos offer a scalable alternative to robot teleoperation for collecting dexterous manipulation data. However, two challenges limit their use as reliable training data: accurate hand-object interaction (HOI) reconstruction from monocular RGB and high-quality robot demonstrations, especially visual observations. We present \textbf{EgoMatrix}, an end-to-end data engine that converts monocular egocentric human videos into bimanual dexterous trajectories and paired robot observation videos, with the trajectories directly replayable on real robots. It reconstructs HOI with emphasis on relative geometry and contact rather than exact metric scene recovery, then uses retargeting and reinforcement learning (RL) to reproduce the demonstrated tasks in simulation. We jointly render robot hands and objects from these simulated trajectories to produce physically faithful observations synchronized with state and control records. Rendering the same trajectories in diverse, task-compatible backgrounds expands paired training data while preserving robot-object motion and contact. Across TACO, OakInk2, and self-collected data, EgoMatrix achieves 54.84\% end-to-end success in simulation versus 14.52\% for the strongest baseline. Reconstruction reaches state-of-the-art performance on most metrics of the two public benchmarks. Qualitative results show more faithful hand-object interactions than prior methods, while feature-space analysis shows closer alignment with real-robot observations than GPT-6 Astra. Our project page is available at \url{https://spikew726.github.io/EgoMatrix/}.
\end{abstract}

\section{Introduction}
\label{sec:intro}

Learning generalizable dexterous manipulation policies requires large, diverse datasets pairing visual observations with robot actions \citep{kim2025openvla}. Teleoperation and motion capture provide high-quality demonstrations, but specialized hardware and operator effort make them costly to scale, especially for multi-fingered hands \citep{wang2024dexcap,xu2025dexumi}. Egocentric videos offer a more accessible source of diverse human manipulation \citep{paliwal2026doasido}, yet converting them into robot demonstrations of comparable quality remains challenging \citep{liu2026egoengine}.

Recent systems have shown that manipulation can be transferred from monocular human videos to dexterous robots \citep{paliwal2026doasido,mu2026deximit,ren2026c2dex}. However, different combinations of 3D scale, depth, and hand-object placement can produce similar 2D projections, making relative geometry and contact difficult to recover reliably from monocular images. A reconstruction may therefore match the input images while still containing hand-object gaps or interpenetration. Such errors can conflict with the contacts required for successful manipulation.

Generating visual observations consistent with the robot's actual manipulation poses a second challenge after successful task transfer in simulation. Data engines such as EgoEngine generate paired robot observations and trajectories \citep{liu2026egoengine}, but retaining the original object imagery can leave depicted contacts and object motion inconsistent with the robot trajectory. For policy learning, visual observations should reflect how the robot actually manipulates the object.

\begin{figure}[!t]
    \centering
    \includegraphics[width=\linewidth]{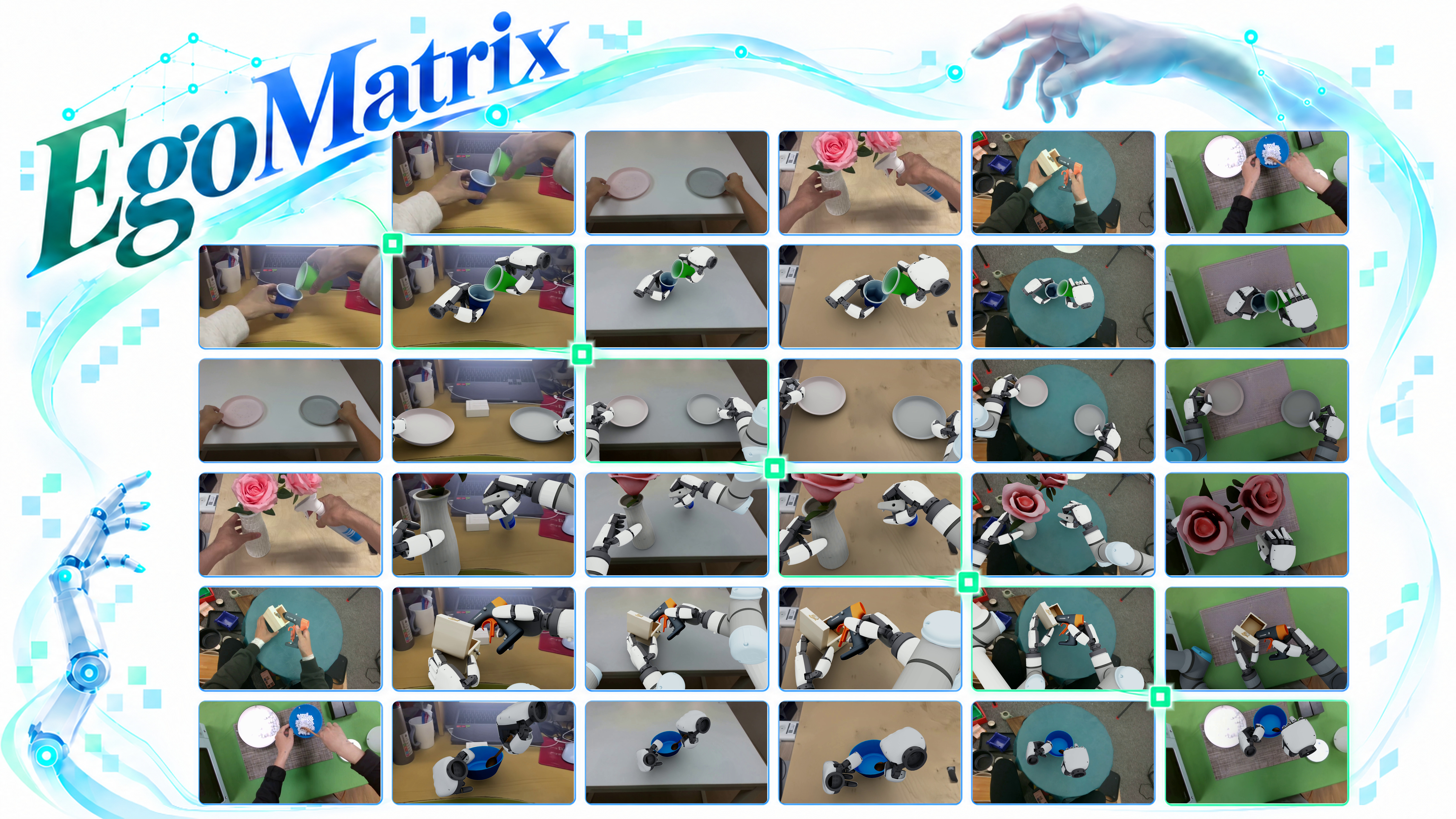}
    \captionsetup{font=small}
    \caption{\textbf{EgoMatrix: Any Task in Any Background.} Rows denote source egocentric human manipulation tasks and columns denote target backgrounds; each cell shows the corresponding dexterous robot demonstration. EgoMatrix preserves physically faithful hand-object motion and contact across these task-background combinations, substantially increasing the visual diversity of robot training data.}
    \label{fig:teaser}
\end{figure}

We present \textbf{EgoMatrix}, an end-to-end data engine that converts monocular egocentric human videos into bimanual dexterous robot trajectories and synchronized observation videos. Our reconstruction prioritizes relative hand-object geometry and contact over exact metric scene recovery. We first use reconstructed hand geometry and relative-depth cues to establish the scene scale and initialize hand-object alignment. Multi-start pose refinement then corrects wrist and object misalignments, followed by local contact refinement to close contact gaps and reduce interpenetration. The refined trajectories guide retargeting and RL, enabling robots to complete the demonstrated tasks in simulation.

We then jointly render the robot hands and objects from successful simulated trajectories to produce physically faithful observations synchronized with state and control records. These observations are composited into the source scene with simulated shadows, while support-plane alignment and target-view rendering extend the same manipulation to diverse, task-compatible backgrounds.

This enables demonstrations from large egocentric video datasets to be converted into paired robot data for a specified real-world workstation, potentially reducing additional robot data collection when adapting pretrained policies to that workstation.

Our contributions are threefold:
\begin{itemize}
\item We present \textbf{EgoMatrix}, an end-to-end data engine that converts monocular egocentric human videos into bimanual dexterous robot demonstrations with synchronized observations and state-control trajectories, supporting direct trajectory replay on real robots. EgoMatrix achieves 54.84\% end-to-end success in simulation versus 14.52\% for the strongest baseline.

\item We develop \textbf{contact-aware hand-object reconstruction} with multi-start pose refinement and local contact refinement, achieving state-of-the-art performance on most reconstruction metrics. Ablations show that both stages improve end-to-end success in simulation.

\item We introduce \textbf{physically faithful robot observation synthesis} by jointly rendering robot hands and objects from successful simulated trajectories. Support-plane alignment and target-view rendering expand the paired data across diverse, task-compatible backgrounds while preserving robot-object motion and contact.
\end{itemize}
\begin{figure}[!t]
    \centering
    \includegraphics[width=\linewidth]{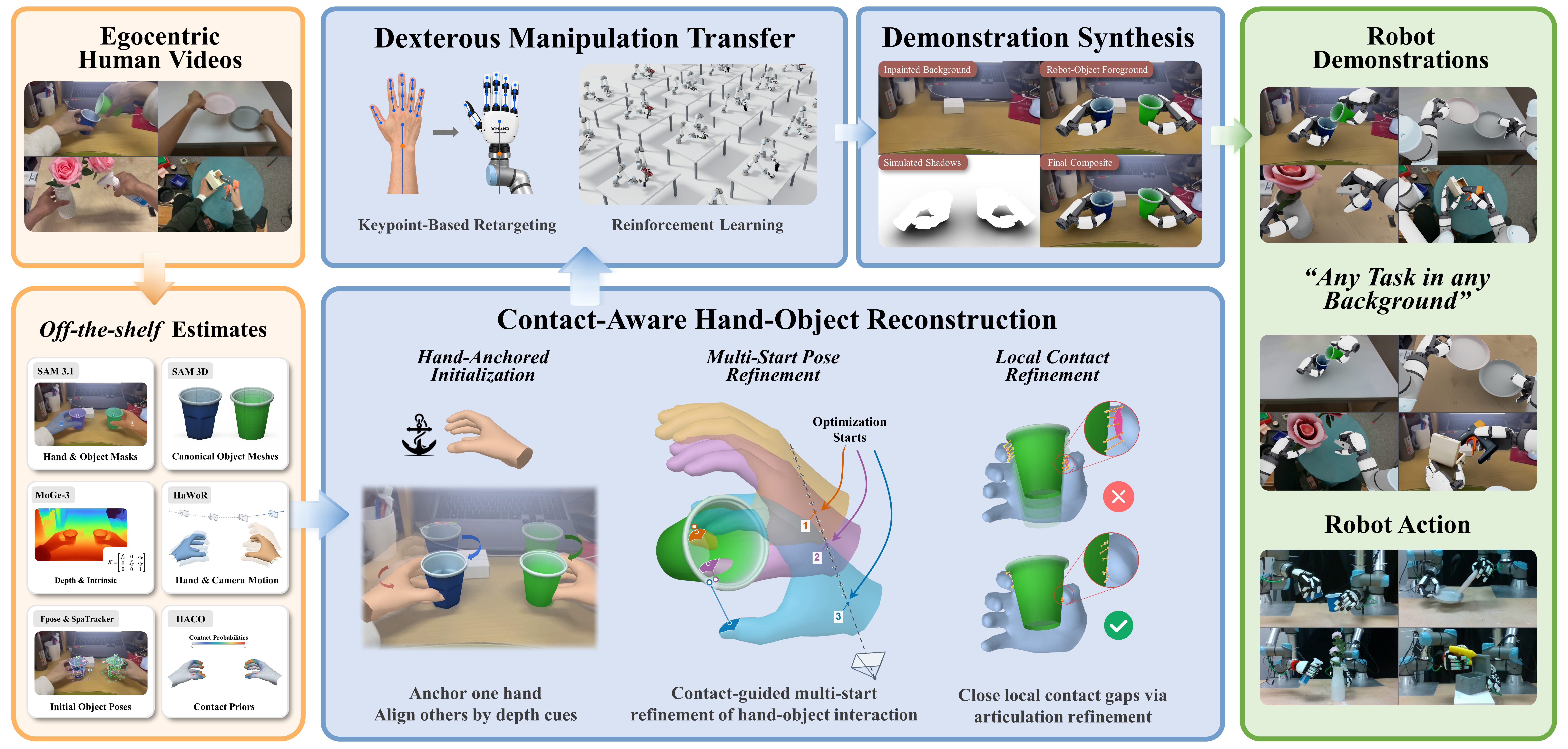}
    \captionsetup{font=small}
    \caption{\textbf{Overview of EgoMatrix.} Starting from monocular egocentric human videos, EgoMatrix performs contact-aware hand-object reconstruction, dexterous manipulation transfer via retargeting and RL, and physically faithful robot observation synthesis. It produces robot trajectories executable on real hardware and synchronized observation videos, with the same trajectories transferable to diverse, task-compatible backgrounds.}
    \label{fig:overview}
\end{figure}

\section{Related Work}
\label{sec:related_work}

\paragraph{Monocular Hand-Object Reconstruction.}
Monocular hand-object reconstruction recovers 3D hand and object geometry from RGB videos, often by matching rendered reconstructions to image-space observations \citep{fan2024hold,aboukhadra2026ghost}. Occlusion limits this evidence, motivating learned shape and generative priors for hidden object regions \citep{wang2025magichoi,chen2026forehoi}. Since geometry completion alone does not resolve ambiguous hand-object placement and contact, contact maps and temporally aggregated correspondences provide additional alignment cues \citep{grady2021contactopt,ren2026c2dex}.  EgoMatrix similarly exploits contact cues, using contact-guided multi-start pose optimization under image-space constraints and subsequent local contact refinement to better preserve the HOI demonstrated in the human videos.

\paragraph{Human-to-Robot Manipulation Transfer.}
After reconstruction, human hand-object motion must be adapted to the robot embodiment while preserving the demonstrated interaction. Existing methods combine geometric retargeting \citep{qin2022dexmv,zhang2026unidex} with grasp and motion planning \citep{mu2026deximit}, reference-tracking reinforcement learning \citep{li2025maniptrans,ren2026c2dex}, or sampling-based trajectory optimization \citep{paliwal2026doasido} to obtain executable robot motion. EgoMatrix follows the reference-tracking approach, using contact-aware reconstructed trajectories as motion references for policy learning.

\paragraph{Robot Observation Synthesis.}
Robot observations can be synthesized by replacing human hands while retaining source object content \citep{zhang2026unidex,liu2026egoengine}, or by generative models conditioned on images or simulated geometry \citep{chen2025roviaug,chen2026craft}. These image-level approaches may leave depicted contacts or object motion inconsistent with the robot trajectory. A different line of work starts from recorded robot demonstrations and uses physics simulation to generate additional training data \citep{jiang2025dexmimicgen,fang2025rebot}. EgoMatrix instead starts from monocular human videos, transfers the demonstrated manipulation to successful robot trajectories in simulation, and jointly renders the robot hands and objects from those trajectories to produce physically faithful observations.

\section{Method}
\label{sec:method}

Given a monocular egocentric human video, EgoMatrix reconstructs hand-object interaction, transfers the demonstrated task to a dexterous robot via retargeting and RL, and synthesizes synchronized robot observations and motion data (Fig.~\ref{fig:overview}). The resulting trajectories can also be rendered in different task-compatible backgrounds.

\subsection{Contact-Aware Hand-Object Reconstruction}
\label{sec:reconstruction}

Without any external calibration or scene-specific metric references, monocular RGB observations cannot uniquely determine metric scene geometry. We therefore target two properties: \textbf{manipulation-relevant 3D consistency} for motion transfer and \textbf{source-view consistency} for video compositing. Our reconstruction proceeds from hand-anchored initialization to multi-start pose refinement and local contact refinement.

\paragraph{Hand-Anchored Initialization}
\label{sec:hand_anchored_init}

We use SAM~3.1 \citep{carion2025sam3segmentconcepts} with natural-language prompts to obtain hand and object masks, and SAM~3D \citep{chen2026sam3d} to reconstruct an unscaled canonical mesh $\mathcal{V}_j$ for each object. The temporal median of the per-frame intrinsics from MoGe-3 \citep{kong2026moge3} defines a sequence-shared camera matrix $\mathbf{K}$, while HaWoR \citep{zhang2025hawor} provides the MANO \citep{MANO:SIGGRAPHASIA:2017} hand trajectories.

We use the human-scale prior encoded by the reconstructed MANO geometry as the scene-scale anchor. Let $\mathcal{P}_{h,t}$ denote pixels shared by the visible hand mask and rendered MANO surface. We estimate a sequence-level scale from the MANO depth $z^{H}_{h,t}(u)$ and MoGe-3 depth $D^{M}_{t}(u)$:

\begin{equation}
\log s_{\mathrm{seq}}
=
\mathcal{R}
\left(
\left\{
\log z^{H}_{h,t}(u)-\log D^{M}_{t}(u)
\right\}_{h,t,u}
\right),
\qquad
\widetilde{D}_{t}
=
s_{\mathrm{seq}}D^{M}_{t},
\label{eq:scale_alignment}
\end{equation}

where $\mathcal{R}$ robustly aggregates observations over $u\in\mathcal{P}_{h,t}$. Back-projecting the aligned depth $\widetilde{D}_{t}$ within each object mask to 3D yields an object point cloud, from which we estimate a sequence-shared uniform scale $s_j$ for $\mathcal{V}_j$. We also use $\widetilde{D}_t$ as the metric-depth input to HaWoR's SLAM module, so that the camera translation is recovered under the same scale.

Using the scaled meshes, aligned depth, and camera trajectory, we initialize object poses with FoundationPose \citep{wen2024foundationpose}, augmented by rigid-motion hypotheses from SpatialTracker \citep{xiao2024spatialtracker} to improve propagation under hand occlusion.

To first reduce large relative-placement errors with a lightweight alignment before multi-start refinement, we fix the right hand and optimize the frame-wise translations of the left hand and both objects, while keeping hand articulation, object rotation, and object scale fixed:

\begin{equation}
\mathcal{L}_{\mathrm{align}}
=
\lambda_{\mathrm{2D}}\mathcal{L}_{\mathrm{2D}}
+
\underbrace{
\lambda_{\mathrm{ord}}\mathcal{L}_{\mathrm{ord}}
+
\lambda_{\mathrm{gap}}\mathcal{L}_{\mathrm{gap}}
}_{\mathcal{L}_{\mathrm{depth}}}
+
\lambda_{\mathrm{temp}}\mathcal{L}_{\mathrm{temp}}
+
\lambda_{\mathrm{prior}}\mathcal{L}_{\mathrm{prior}}.
\label{eq:init_alignment}
\end{equation}

Here, $\mathcal{L}_{\mathrm{2D}}$ encourages source-view consistency through visibility and silhouette constraints, while $\mathcal{L}_{\mathrm{depth}}$ preserves pairwise front-back ordering and relative depth differences induced by $\widetilde{D}_t$; the remaining terms regularize the trajectory. 

We estimate contact frames for each hand using HACO~\citep{jung2026learning} and the proximity of the observed hand and object masks. Specifically, we project MANO vertices into the source image and use their HACO contact probabilities when they lie near both the hand mask and the corresponding object mask. We combine this evidence with the distance between the two masks, then apply temporal filtering to identify sustained contact intervals. Let $m_{h,t}\in\{0,1\}$ denote the resulting contact mask. We define
\begin{equation}
\mathcal{T}_h^{\mathrm{c}}
=
\{t\in\mathcal{T}_h\mid m_{h,t}=1\},
\qquad h\in\{L,R\},
\label{eq:contact_frames}
\end{equation}
where $\mathcal{T}_h$ denotes the task frames for hand $h$. The two hands are processed independently.

We additionally estimate gravity and a shared support surface from the background geometry. Frames outside $\mathcal{T}_h^{\mathrm{c}}$ provide evidence for stable object support, while objects are not forced onto the support surface during contact. Further details are provided in Appendix~\ref{app:hand_anchored_init}.

\Needspace{18\baselineskip}

\begin{wrapfigure}[18]{r}{0.30\textwidth}
    \centering
    \includegraphics[width=\linewidth]{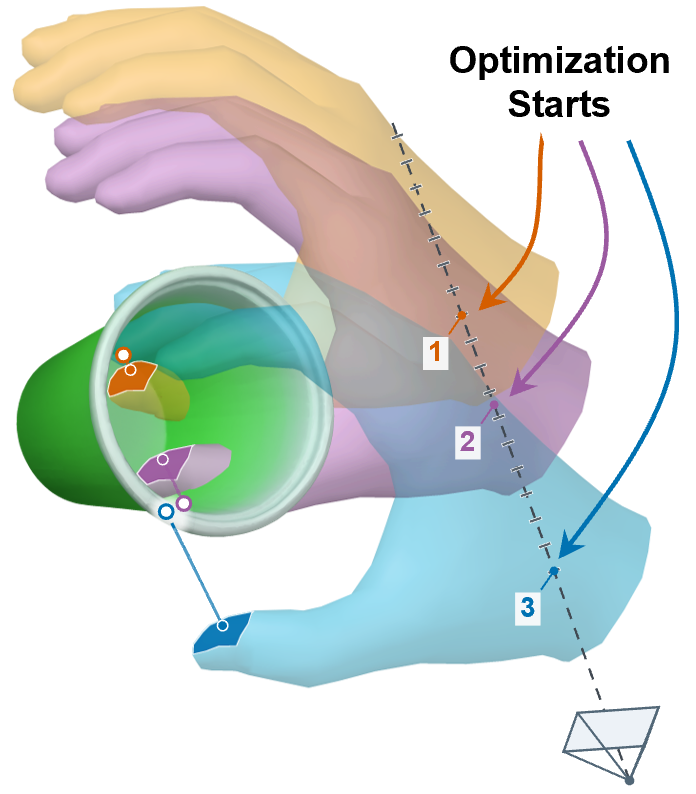}
    \captionsetup{
        font=small,
        skip=2pt,
        justification=raggedright,
        singlelinecheck=false
    }
    \caption{
    \textbf{Contact-guided multi-start initialization.}
    Different hand offsets associate candidate hand vertices with different points on the object surface, providing alternative starts for pose refinement.
    }
    \label{fig:multistart_initialization}
    \vspace{-0.2\baselineskip}
\end{wrapfigure}

\paragraph{Multi-Start Pose Refinement}
\label{sec:coarse_refinement}

After the hand-anchored initialization stage, we refine the wrist and object rigid poses while keeping MANO articulation fixed. For contact frames $t\in\mathcal{T}_h^{\mathrm{c}}$, HACO \citep{jung2026learning} provides contact probabilities $p_{h,t,i}$, from which we define likely contact vertices $\mathcal{I}_{h,t}=\{i\mid p_{h,t,i}>\tau_{\mathrm{H}}\}$. For each MANO vertex $i \in \mathcal{I}_{h,t}$, we consider the camera ray passing through it. On concave or multi-layered objects, the ray may intersect multiple surfaces, and normal compatibility alone may still leave several plausible intersections on the object surface for the same hand vertex. Since pose refinement is sensitive to initialization \citep{grady2021contactopt}, we construct multiple contact-guided starts that cover these alternative surface associations.

As illustrated in Fig.~\ref{fig:multistart_initialization}, we generate starting poses by translating the hand along the ray from the camera center through the mean position of the initial MANO vertices. For each candidate $k$, we cast rays through the vertices in $\mathcal{I}_{h,t}$ and select the intersection closest to the corresponding vertex. These intersections are stored in object coordinates and define the targets of $\mathcal{L}_{\mathrm{ray}}^{(k)}$, which attracts the hand vertices toward them during subsequent pose optimization.

\WFclear

To compare candidates, we group vertices in $\mathcal{I}_{h,t}$ into fixed patches using MANO mesh connectivity. For each patch, we record how often vertices are paired with the first, second, or later intersection along the ray. For each candidate, we compute its difference from every retained start and use the smallest difference as its diversity measure. Starting from the unshifted pose, we greedily select additional candidates with large diversity while penalizing large distances between paired points. For task frames outside $\mathcal{T}_h^{\mathrm{c}}$, we use only the unshifted pose.

We then jointly refine the wrist and bound-object poses from each selected start. We collect the source-view, relative-depth, penetration, and regularization terms into $\mathcal{L}_{\mathrm{base}}$. The objective is

\begin{equation}
\mathcal{L}_{\mathrm{coarse}}^{(k,s)}
=
\mathcal{L}_{\mathrm{base}}
+
\lambda_{\mathrm{c}}
\left[
\eta_s\mathcal{L}_{\mathrm{ray}}^{(k)}
+
(1-\eta_s)\mathcal{L}_{\mathrm{surf}}
\right],
\label{eq:coarse_objective}
\end{equation}

where $s$ indexes optimization steps and $\mathcal{L}_{\mathrm{surf}}$ attracts vertices in $\mathcal{I}_{h,t}$ toward the current nearest object surface. The weight $\eta_s$ decreases from $1$ to $0$, so all starts are ultimately refined under the same contact objective. The contact attraction terms apply only to frames in $\mathcal{T}_h^{\mathrm{c}}$. For other task frames, optimization relies on the remaining terms in $\mathcal{L}_{\mathrm{base}}$. Candidate selection and optimization details are provided in Appendix~\ref{app:multi_start_refinement}.

\paragraph{Local Contact Refinement}
\label{sec:fine_refinement}

To improve the local hand-object geometry used for downstream manipulation transfer, we finally fix the object poses and jointly refine the wrist pose and MANO articulation. For frames $t\in\mathcal{T}_h^{\mathrm{c}}$, we cast rays through temporally filtered HACO contact vertices onto the object. When multiple intersections occur, we select those close to the hand vertex in depth with the strongest opposing normal match. These frame-wise contacts are stabilized in canonical object space following C2Dex \citep{ren2026c2dex}. When stable patches provide insufficient opposing support, we supplement region-aware contacts according to the pad, palm, and side opposition types of the GRASP taxonomy \citep{feix2016grasp}.

For stable contact vertex $i$, let $\mathbf{v}_{t,i}$ and
$\mathbf{y}^{*}_{t,i}$ denote the current hand vertex and its
object-side target. For concise notation, we write the contact attraction as
\begin{equation}
\mathcal{L}_{\mathrm{contact}}
=
\frac{1}{|\mathcal{T}_h^{\mathrm{c}}|}
\sum_{t\in\mathcal{T}_h^{\mathrm{c}}}
\sum_{i\in\mathcal{C}_{t}}
w_{t,i}
\left\|
\mathbf{v}_{t,i}
-
\mathbf{y}^{*}_{t,i}
\right\|_{2}^{2},
\label{eq:fine_contact}
\end{equation}

where $\mathcal{C}_{t}$ contains vertices with valid contact targets, including supplementary opposition contacts when needed, and $w_{t,i}$ is a contact weight derived from HACO confidence and temporal support. We combine $\mathcal{L}_{\mathrm{contact}}$ with bidirectional penetration penalties and regularization terms to form the final contact refinement objective. For frames outside $\mathcal{T}_h^{\mathrm{c}}$, wrist pose and MANO articulation are still optimized using penetration penalties and regularization, without contact attraction.Appendix~\ref{app:local_contact_refinement} provides the complete contact construction and optimization details.

\begin{figure}[!t]
    \centering
    \includegraphics[width=0.94\linewidth]{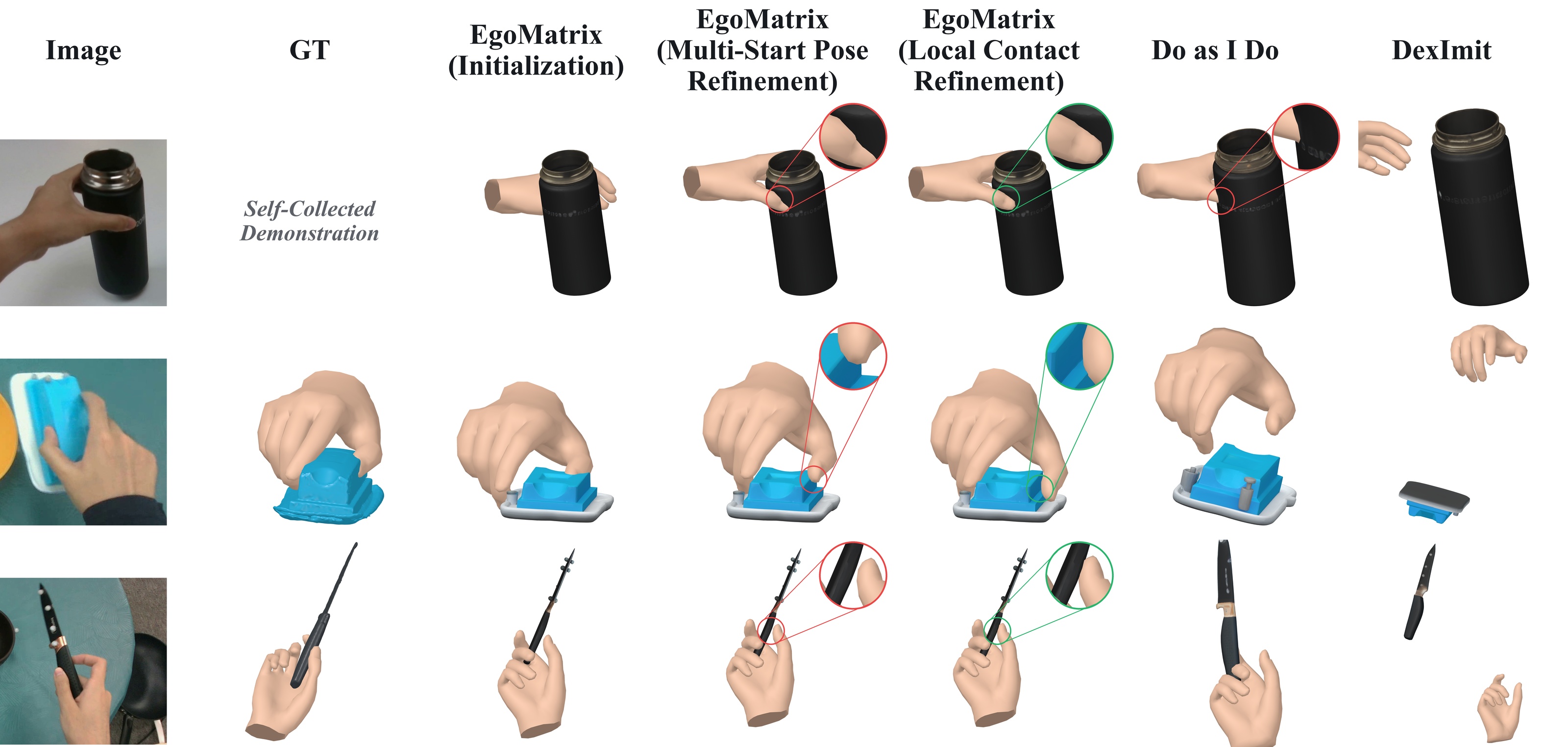}

    \captionsetup{font=small}
    \caption{
    \textbf{Qualitative comparison of hand-object reconstruction.}
    Each row shows the same input frame across ground truth, the three stages of EgoMatrix, and competing methods.
    The three examples illustrate how Local Contact Refinement addresses three contact defects that can hinder downstream dexterous-hand transfer: \textbf{severe interpenetration} (top), \textbf{poorly localized contact} (middle), and \textbf{residual hand-object gaps} (bottom).
    Across these examples, EgoMatrix produces more interaction-consistent hand-object reconstructions than the competing methods.
    }
    \label{fig:reconstruction_comparison}
\end{figure}

\subsection{Dexterous Manipulation Transfer}
\label{sec:motion_transfer}

Given the reconstructed hand-object trajectories, we deliberately use a lightweight keypoint-based retargeting scheme rather than a more elaborate retargeting method. For each hand, we define fixed semantic correspondences between MANO joints
and fingertips and robot-hand links. Let $\mathbf{x}_{t,\ell}\in\mathbb{R}^3$
denote the 3D position of MANO keypoint $\ell$ at frame $t$, and let
$\mathbf{p}_t$, $\mathbf{R}_t$, and $\mathbf{q}_t$ denote the robot wrist
position, wrist orientation, and finger configuration. With
$\mathbf{f}_{\ell}(\mathbf{q}_t)$ denoting the wrist-local forward-kinematics
position of the corresponding robot keypoint, we optimize

\begin{equation}
\mathcal{L}_{\mathrm{ret}}
=
\frac{1}{TN}
\sum_{t=1}^{T}
\sum_{\ell=1}^{N}
w_{\ell}
\left\|
\mathbf{p}_t
+
\mathbf{R}_t\mathbf{f}_{\ell}(\mathbf{q}_t)
-
\mathbf{x}_{t,\ell}
\right\|_2
+
\mathcal{R}_{\mathrm{ret}}.
\end{equation}

subject to the robot joint limits. $\mathcal{R}_{\mathrm{ret}}$ regularizes wrist orientation, temporal smoothness, and table clearance. The two hands are retargeted independently, and the resulting wrist and joint trajectories are used to initialize physics-based policy learning.

We then adopt the RL framework of ManipTrans \citep{li2025maniptrans}, but omit its pretrained imitator and train a joint bimanual policy directly in Isaac~Lab~\citep{mittal2025isaaclab}. The retargeted robot trajectory is used for reference-state initialization, while the reconstructed MANO and object trajectories serve as tracking references. We train the policy with PPO~\citep{schulman2017proximal}, using rewards for hand and object tracking, contact, and power. The learned policy is then rolled out in simulation to obtain a physically executable bimanual trajectory.

\subsection{Demonstration Synthesis and Background Transfer}
\label{sec:demo_synthesis}

Following a successful policy rollout, we replay the recorded simulator states and jointly render the dexterous hands and reconstructed objects using the source-video camera intrinsics and frame-aligned camera poses. Because both the hands and objects are rendered from the same physics rollout, their relative geometry directly reflects the executed interaction. We remove the human hands, arms, and manipulated objects from the source video using SVOR \citep{hu2026svor} with removal masks derived from SAM~3.1 segmentation. We then estimate tabletop shadow attenuation from paired simulator renders with and without the robot-object foreground, and apply it to the cleaned background. 

Let $I_t^{\mathrm{sim}}$ denote the simulator-rendered RGB image, $M_t^{\mathrm{sim}}$ the combined binary mask of the robot hands and manipulated objects, and $\widetilde{I}_t^{\mathrm{bg}}$ the shadow-adjusted background. The synthesized frame is
\begin{equation}
I_t^{\mathrm{syn}}
=
M_t^{\mathrm{sim}}\odot I_t^{\mathrm{sim}}
+
\left(1-M_t^{\mathrm{sim}}\right)\odot
\widetilde{I}_t^{\mathrm{bg}}.
\label{eq:demo_compositing}
\end{equation}

The same successful execution can be transferred to another task-compatible background without modifying its internal hand-object geometry. For source and target task frames $\mathcal{F}_A$ and $\mathcal{F}_B$, respectively defined from the support surface and initial object placement of each scene, we compute a single rigid transform $\mathbf{T}_{B\leftarrow A}\in\mathrm{SE}(3)$ and apply it to the entire interaction. Thus, both hands and all manipulated objects preserve their relative poses, with no rescaling or frame-wise realignment. We re-render the same robot and object trajectories using the target camera trajectory and support-surface proxy. The rendered foreground is then composited with the cleaned target background using the same synthesis procedure. Different compatible backgrounds can therefore reuse the same robot and object trajectories and control records, changing only the common rigid transform applied to the trajectories, the camera viewpoint, and the visual background. Figure~\ref{fig:teaser} illustrates this transfer across multiple source manipulation tasks and target backgrounds. Details of scene alignment and temporal synchronization are provided in Appendix~\ref{app:background_transfer}.
\section{Experiments}
\label{sec:exp}

\subsection{Experimental Setup}
\label{sec:experimental_setup}

\noindent\textbf{Datasets.}
We evaluate on TACO~\citep{liu2024taco}, OakInk2~\citep{zhan2024oakink2}, and a self-collected dataset. We use $4$ TACO tasks with $37$ demonstrations, $2$ OakInk2 tasks with $14$ demonstrations, and $11$ self-collected tasks with one demonstration each. TACO and OakInk2 are used for reconstruction evaluation; all three datasets are used for end-to-end manipulation transfer.

\vspace{2pt}
\noindent\textbf{Baselines.}
For end-to-end transfer, we compare with DexImit~\citep{mu2026deximit} and Do As I Do~\citep{paliwal2026doasido}. For reconstruction, we compare with G-HOP~\citep{ye2024ghop}, GHOST~\citep{aboukhadra2026ghost}, MagicHOI~\citep{wang2025magichoi}, and ForeHOI~\citep{chen2026forehoi}. For visual synthesis, we compare qualitatively with EgoEngine~\citep{liu2026egoengine}.

\vspace{2pt}
\noindent\textbf{Metrics.}
For HOI reconstruction, we report root-aligned mean per-joint position error (MPJPE$_{\mathrm{RA}}$) for hand pose; Chamfer distance (CD) and F-score at a $5$\,mm threshold (F@5) for object geometry; absolute trajectory error (ATE), absolute rotation error (ARE), relative translation error (RTE), and relative rotation error (RRE) for object trajectories~\citep{shin2026compose}; and hand-relative Chamfer distance (CD$_h$), simulation displacement (SD), and maximum penetration depth ($D_{\mathrm{pen}}^{\max}$) for interaction quality~\citep{fan2024hold,jiang2021grasptta,hasson2019joint}. We also report the \textit{Trajectory Transfer Success Rate}, defined as the fraction of demonstrations converted into robot trajectories that complete the same task in simulation. For TACO and OakInk2, we align the simulated and ground-truth object trajectories as in ATE and count a transfer as successful if the translation and rotation errors remain within $0.10$\,m and $1.0$\,rad throughout the task. For self-collected videos, where ground-truth object poses are unavailable, we manually judge each rollout and require both a valid grasp and completion of the intended manipulation.

\subsection{End-to-End Video-to-Robot Trajectory Transfer}
\begin{table}[t]
    \centering
    \caption{
        End-to-end success rate of trajectory transfer across TACO, OakInk2, and self-collected demonstrations. Each entry reports the success rate and the number of successful trajectories over the total number of evaluated demonstrations.
    }
    \label{tab:trajectory_success}

    \small
    \setlength{\tabcolsep}{6pt}
    \renewcommand{\arraystretch}{1.0}

    \begin{tabular}{llccc}
        \toprule

        \multirow{2}{*}{\textbf{Dataset}}
        & \multirow{2}{*}{\textbf{Tasks}}
        & \multicolumn{3}{c}{\textbf{Trajectory Transfer Success Rate (\%) $\uparrow$}} \\

        \cmidrule(lr){3-5}

        &
        & \textbf{DexImit}\textsuperscript{*}
        & \textbf{Do As I Do}
        & \textbf{EgoMatrix} \\

        \midrule

        \multirow{4}{*}{TACO}
        & brush, brush, bowl
        & 7.14\% (0.5/7)
        & 28.57\% (2/7)
        & \textbf{71.43\% (5/7)} \\

        & cut, knife, bowl
        & 20.83\% (2.5/12)
        & 0.00\% (0/12)
        & \textbf{33.33\% (4/12)} \\

        & screw, screwdriver, box
        & 0.00\% (0/11)
        & 9.09\% (1/11)
        & \textbf{45.45\% (5/11)} \\

        & smear, glue gun, box
        & 0.00\% (0/7)
        & 0.00\% (0/7)
        & \textbf{57.14\% (4/7)} \\

        \midrule

        \multirow{2}{*}{OakInk2}
        & stir spoon/rod
        & 0.00\% (0/8)
        & 25.00\% (2/8)
        & \textbf{62.50\% (5/8)} \\

        & rearrange tableware
        & 0.00\% (0/6)
        & 16.67\% (1/6)
        & \textbf{50.00\% (3/6)} \\

        \midrule

        Self-Collected
        & All 11 tasks
        & 45.45\% (5/11)
        & 27.27\% (3/11)
        & \textbf{72.73\% (8/11)} \\

        \midrule

        \textbf{Overall}
        & \textbf{All above tasks}
        & 12.90\% (8/62)
        & 14.52\% (9/62)
        & \textbf{54.84\% (34/62)} \\

        \bottomrule
    \end{tabular}

    \vspace{2pt}
\begin{minipage}{\linewidth}
\footnotesize
\textsuperscript{*}For DexImit, a demonstration receives $0.5$ credit when only one of the two hand-object trajectories satisfies the success criterion. Requiring both hands to succeed would give DexImit $0\%$ on all public-dataset tasks. 
\end{minipage}
\end{table}

First, we evaluate end-to-end trajectory transfer on selected TACO and OakInk2 tasks and 11 self-collected demonstrations spanning diverse object geometries, grasp classes, and motion patterns. As shown in Table~\ref{tab:trajectory_success}, EgoMatrix achieves the highest success rate across all evaluated task groups, reaching 54.84\% (34/62) overall, compared with 14.52\% (9/62) for Do As I Do and 12.90\% (8/62) for DexImit. The gains are consistent across all three data sources and include tasks where the baselines rarely or never succeed, such as \textit{screw, screwdriver, box} and \textit{smear, glue gun, box}. Overall, EgoMatrix converts monocular human demonstrations into robot trajectories that reproduce the demonstrated manipulation with substantially higher success rates.

\subsection{Reconstruction Quality}
\label{sec:reconstruction_results}

We next evaluate reconstruction quality, with particular attention to the hand-object interaction used for downstream transfer. As shown in Table~\ref{tab:reconstruction_comparison}, EgoMatrix achieves the best result on $8$ of $10$ metrics on TACO and $6$ of $10$ on OakInk2, including the best CD, F@5, ATE, and RTE on both datasets. It also achieves the best CD$_h$ and SD on TACO and remains competitive on OakInk2.

Interaction metrics should be interpreted jointly. DexImit, for example, obtains the lowest $D_{\mathrm{pen}}^{\max}$ on both datasets but substantially worse CD$_h$ and SD, showing that low penetration alone does not imply accurate hand-object interaction. Fig.~\ref{fig:reconstruction_comparison} further shows more interaction-consistent reconstructions than Do As I Do and DexImit on representative examples.

\begin{table}[t]
    \centering
    \setlength{\belowcaptionskip}{5pt}
    \captionsetup{font=small}
    \caption{
        \textbf{Quantitative comparison of hand-object reconstruction on TACO and OakInk2.} Lower is better except for F@5. Best results are shown in bold and second-best results are underlined. ``--'' denotes metrics unavailable from the method outputs.
    }
    \label{tab:reconstruction_comparison}

    \small
    \setlength{\tabcolsep}{1.35pt}
    \renewcommand{\arraystretch}{1.03}

    \begin{tabular}{@{}ll@{\hspace{3pt}}c@{\hspace{-3pt}}cccccccccc@{}}
        \toprule

        \multirow{3}{*}{{\footnotesize\textbf{Dataset}}}
        & \multirow{3}{*}{{\footnotesize\textbf{Method}}}
        & \multicolumn{2}{c}{{\footnotesize\textbf{Object Geometry}}}
        & \multicolumn{1}{c}{{\footnotesize\textbf{Hand Pose}}}
        & \multicolumn{4}{c}{{\footnotesize\textbf{Object Trajectory}}}
        & \multicolumn{3}{c}{{\footnotesize\textbf{Interaction}}} \\

        \cmidrule(lr){3-4}
        \cmidrule(lr){5-5}
        \cmidrule(lr){6-9}
        \cmidrule(lr){10-12}

        &
        & {\footnotesize\textbf{CD} $\downarrow$}
        & {\footnotesize\textbf{F@5} $\uparrow$}
        & {\footnotesize\textbf{MPJPE$_{\mathrm{RA}}$} $\downarrow$}
        & {\footnotesize\textbf{ATE} $\downarrow$}
        & {\footnotesize\textbf{ARE} $\downarrow$}
        & {\footnotesize\textbf{RTE} $\downarrow$}
        & {\footnotesize\textbf{RRE} $\downarrow$}
        & {\footnotesize\textbf{CD$_h$} $\downarrow$}
        & {\footnotesize\textbf{SD} $\downarrow$}
        & {\footnotesize$\mathbf{D_{\mathrm{pen}}^{\max}}$ $\downarrow$}
        \\[-2pt]

        &
        & {\scriptsize (cm)}
        & {\scriptsize (\%)}
        & {\scriptsize (mm)}
        & {\scriptsize (cm)}
        & {\scriptsize ($^\circ$)}
        & {\scriptsize (cm)}
        & {\scriptsize ($^\circ$)}
        & {\scriptsize (cm)}
        & {\scriptsize (cm)}
        & {\scriptsize (mm)}
        \\

        \midrule

        \multirow{7}{*}{TACO}
        & {\footnotesize G-HOP}
        & 2.54 & 30.87 & 37.28 & 38.35 & 65.32 & 7.69 & 8.62
        & 37.66 & 9.23 & \second{1.84} \\

        & {\footnotesize MagicHOI}
        & 2.11 & 40.44 & 39.90 & -- & -- & -- & --
        & 13.33 & 9.14 & 17.52 \\

        & {\footnotesize ForeHOI}
        & 1.52 & 51.28 & -- & 16.11 & 69.14 & 2.33 & 8.26
        & -- & -- & -- \\

        & {\footnotesize GHOST}
        & 2.16 & 44.32 & 37.05 & 16.01 & 42.45 & 1.60 & \second{4.20}
        & 23.60 & 5.42 & 17.87 \\

        \cmidrule(lr){2-12}

        & {\footnotesize DexImit}
        & 0.64 & 83.08 & 29.12 & 8.43 & \second{37.53}
        & \second{1.35} & 9.26 & 17.35 & 10.39 & \best{0.35} \\

        & {\footnotesize Do As I Do}
        & \second{0.61} & \second{84.81} & \best{22.16}
        & \second{4.99} & 50.24 & 1.42 & 6.19
        & \second{6.00} & \second{4.63} & 3.69 \\

        & {\footnotesize\textbf{EgoMatrix}}
        & \best{0.56} & \best{86.27} & \second{23.07}
        & \best{3.35} & \best{24.54} & \best{0.70} & \best{3.35}
        & \best{4.06} & \best{2.62} & 2.64 \\

        \midrule

        \multirow{7}{*}{OakInk2}
        & {\footnotesize G-HOP}
        & 1.55 & 42.36 & 27.39 & 46.08 & 63.19 & 3.44 & 6.79
        & \best{11.63} & 9.51 & \second{0.45} \\

        & {\footnotesize MagicHOI}
        & 1.95 & 37.59 & \second{23.52} & -- & -- & -- & --
        & 16.85 & \best{1.94} & 15.71 \\

        & {\footnotesize ForeHOI}
        & 1.00 & 58.36 & -- & 9.31 & 83.09 & 1.43 & 7.13
        & -- & -- & -- \\

        & {\footnotesize GHOST}
        & 1.61 & 50.78 & 29.13 & \second{8.09} & \best{45.44}
        & \second{1.16} & \second{4.00}
        & 18.63 & 2.54 & 21.18 \\

        \cmidrule(lr){2-12}

        & {\footnotesize DexImit}
        & 1.20 & 71.47 & 25.27 & 12.55 & 78.75 & 1.57 & 10.66
        & 30.71 & 9.66 & \best{0.18} \\

        & {\footnotesize Do As I Do}
        & \second{0.59} & \second{84.61} & 24.04
        & 8.93 & 70.19 & 1.79 & 5.19
        & 16.52 & 4.91 & 1.70 \\

        & {\footnotesize\textbf{EgoMatrix}}
        & \best{0.57} & \best{87.75} & \best{20.47}
        & \best{2.52} & \second{58.68} & \best{0.56} & \best{2.70}
        & \second{14.10} & \second{2.52} & 1.81 \\

        \bottomrule
    \end{tabular}
\end{table}

\subsection{Visual Fidelity of Generated Robot Observations}

\begin{figure}[t]
    \centering
    \includegraphics[width=0.94\linewidth]{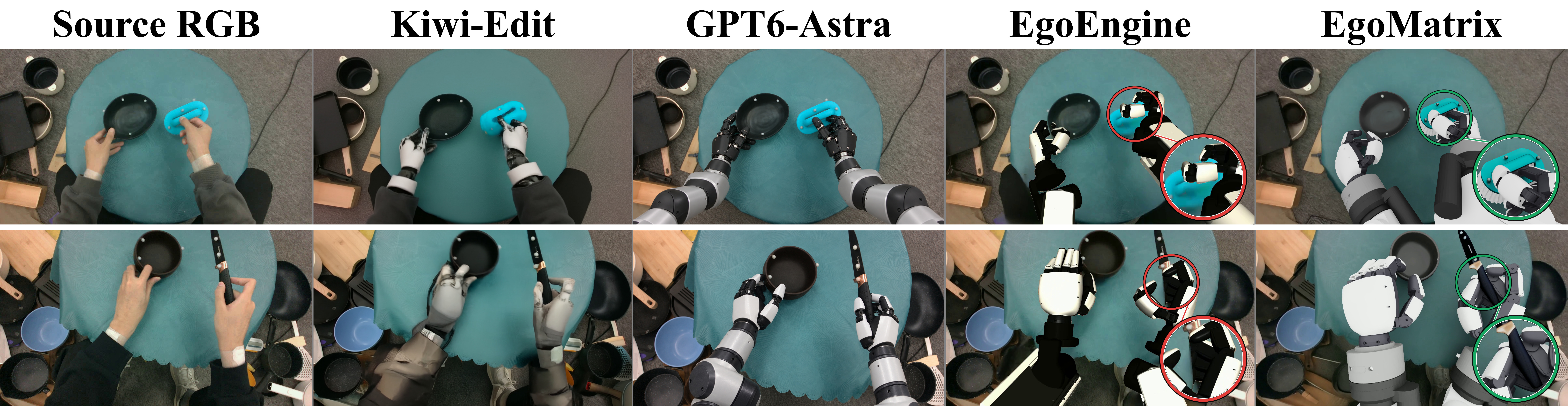}

    \captionsetup{font=small}
    \caption{
    \textbf{Qualitative comparison of robot observation synthesis in the source background.}
    Each row compares Kiwi-Edit, GPT-6 Astra, EgoEngine, and EgoMatrix for the same source frame. Magnified regions highlight representative hand-object contacts. EgoMatrix produces more physically faithful robot observations with hand-object geometry consistent with the transferred manipulation.
    }
    \label{fig:visual_fidelity_source}
\end{figure}

\paragraph{Source-Background Fidelity.}
Fig.~\ref{fig:visual_fidelity_source} compares robot observations synthesized in the original source background. Kiwi-Edit~\citep{lin2026kiwiedit} largely preserves the scene and object placement but often fails to replace the human hands and arms with coherent robot geometry. Despite explicitly specifying a dual-UR5e system with XHand dexterous hands, GPT-6 Astra~\citep{openai2026gpt6astra} still produces image distortion and mismatched robot hands or arms. EgoEngine better preserves robot appearance but exhibits local hand-object contact errors. EgoMatrix instead renders the robot hands and objects from the same successful simulation rollout, preserving the transferred interaction geometry while retaining the source background and viewpoint.

\Needspace{12\baselineskip}

\begin{wraptable}{r}{0.55\textwidth}
    \vspace{-0.9\intextsep}
    \centering
    \captionsetup{font=small,skip=5pt}

    \caption{
        \textbf{Feature alignment (FD) with real robot observations.}
        FD over $52$ frames per domain; lower is better.
    }
    \label{tab:background_fidelity}

    \footnotesize
    \setlength{\tabcolsep}{3.5pt}
    \renewcommand{\arraystretch}{1.1}

    \begin{tabular}{@{}lrrrr@{}}
        \toprule

        \multirow{2}{*}{\textbf{Source}}
        & \multicolumn{4}{c}{\textbf{FD} $\downarrow$} \\

        \cmidrule(lr){2-5}

        & \textbf{ResNet18}
        & \textbf{VGG16}
        & \textbf{DINOv2}
        & \textbf{DreamSim} \\

        \midrule

        Human
        & 273.10 & 1003.64 & 468.06 & 1.0178 \\

        GPT-6 Astra
        & 119.81 & 711.24 & \textbf{240.54} & 0.5934 \\

        EgoMatrix
        & \textbf{100.23} & \textbf{414.66}
        & 253.15 & \textbf{0.4975} \\

        \bottomrule
    \end{tabular}

\end{wraptable}

\begin{figure}[H]
    \centering
    \includegraphics[width=0.94\linewidth]{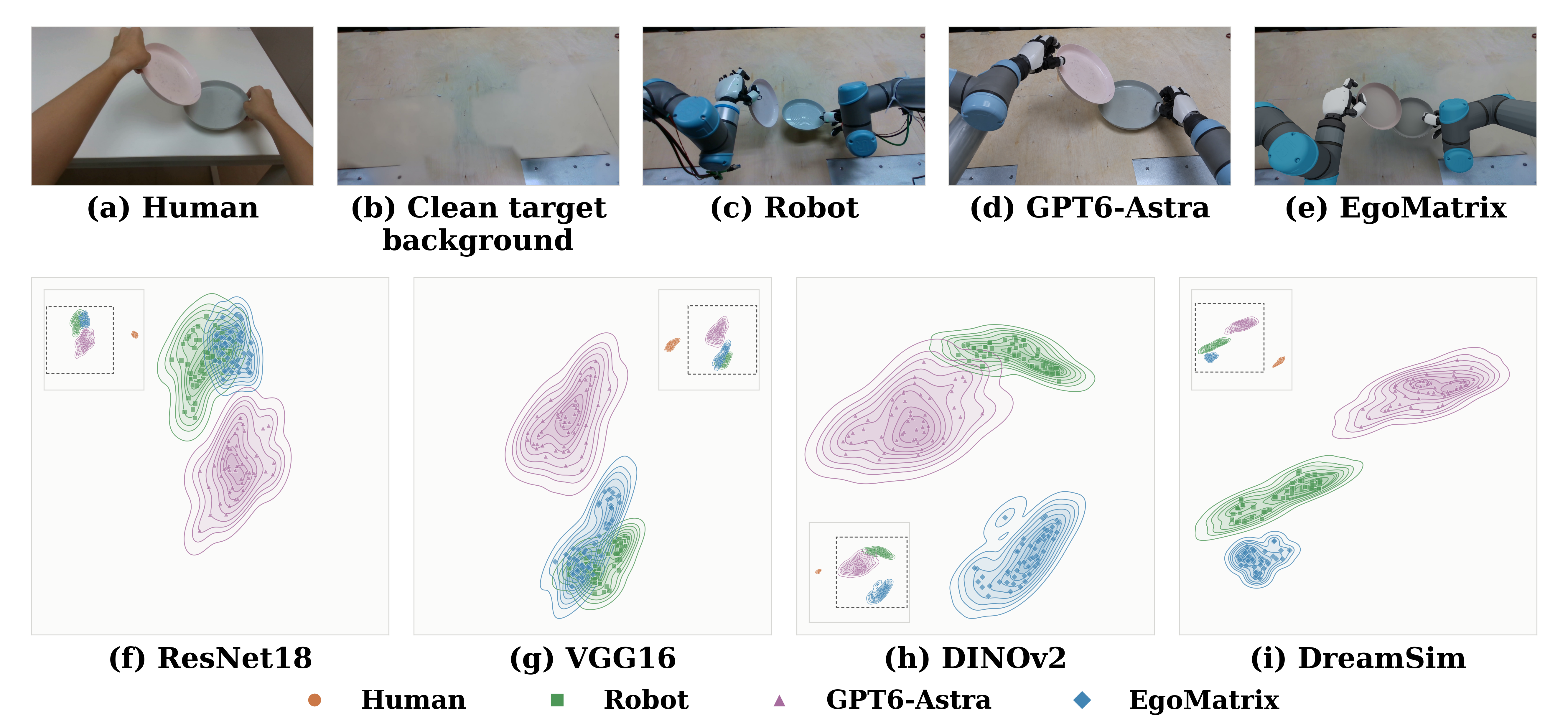}

    \captionsetup{font=small}
    \caption{
    \textbf{Feature-space comparison for target-background transfer.}
    Top: source human frame, clean target background, real-robot observation, GPT-6 Astra, and EgoMatrix.
    Bottom: feature distributions from ResNet18, VGG16, DINOv2, and DreamSim over $52$ frames per domain.
    EgoMatrix more closely matches the real-robot distribution in three of the four feature spaces.
    }
    \label{fig:background_fidelity}
\end{figure}

\paragraph{Target-Background Fidelity.}
We further compare GPT-6 Astra and EgoMatrix against ego-view real-robot observations in a target workspace. Using $52$ frames from one manipulation sequence, we extract features with ResNet18~\citep{he2016deep}, VGG16~\citep{simonyan2015very}, DINOv2~\citep{oquab2024dinov2}, and DreamSim~\citep{fu2023dreamsim}. Fig.~\ref{fig:background_fidelity} visualizes the feature distributions, while Table~\ref{tab:background_fidelity} reports Fr\'echet distance (FD) to the real-robot distribution in each full feature space; lower is better. EgoMatrix obtains the lowest FD for ResNet18, VGG16, and DreamSim, while GPT-6 Astra is slightly better on DINOv2. Qualitatively, Astra often preserves the source object's image-space size and placement under the viewpoint change and fails to reproduce the correct bimanual robot configuration. EgoMatrix re-renders the transferred robot and object trajectories from the target camera, explicitly adapting them to the new viewpoint. The fixed Astra prompt and generation protocol are provided in Appendix~\ref{app:visual_baselines}.

\subsection{Ablation studies}
\label{sec:ablation}

\begin{table}[htbp]
    \centering
    \captionsetup{
        font=small
    }

    \caption{
        \textbf{Ablation of reconstruction refinement.}
        Each row cumulatively adds one stage. We report transfer success on self-collected tasks and TACO, and hand-pose and interaction metrics on TACO.
    }
    \label{tab:ablation_optimization}

    \footnotesize
    \setlength{\tabcolsep}{7pt}
    \renewcommand{\arraystretch}{1.18}

    \begin{tabular}{@{}lcccccc@{}}
        \toprule

        \multirow{3}{*}{\textbf{Method}}
        & \multicolumn{1}{c}{\textbf{Self}}
        & \multicolumn{5}{c}{\textbf{TACO}} \\

        \cmidrule(lr){2-2}
        \cmidrule(lr){3-7}

        & \textbf{Succ.} $\uparrow$
        & \textbf{Succ.} $\uparrow$
        & \textbf{MPJPE$_{\mathrm{RA}}$} $\downarrow$
        & \textbf{CD$_h$} $\downarrow$
        & \textbf{SD} $\downarrow$
        & $\mathbf{D_{\mathrm{pen}}^{\max}}$ $\downarrow$ \\

        & \scriptsize{(\%)}
        & \scriptsize{(\%)}
        & \scriptsize{(mm)}
        & \scriptsize{(cm)}
        & \scriptsize{(cm)}
        & \scriptsize{(mm)} \\

        \midrule

        Hand-Anchored Init.
        & 18.18 & 5.41 & \textbf{21.97}
        & 7.14 & 4.00 & 3.82 \\

        $+$ Multi-Start Refinement
        & 36.36 & 27.03 & 23.17
        & 4.24 & 3.31 & \textbf{1.51} \\

        $+$ Local Contact Refinement
        & \textbf{72.73} & \textbf{48.65}
        & 23.07 & \textbf{4.06}
        & \textbf{2.62} & 2.64 \\

        \bottomrule
    \end{tabular}
\end{table}

Table~\ref{tab:ablation_optimization} shows that both refinement stages are important for downstream manipulation transfer. The hand-anchored initialization already achieves an MPJPE$_{\mathrm{RA}}$ of $21.97$\,mm on TACO, outperforming all baselines in Table~\ref{tab:reconstruction_comparison}. The subsequent refinements slightly degrade this hand-pose metric but substantially improve interaction quality and end-to-end success: on TACO, success rises from $5.41\%$ to $27.03\%$ and then $48.65\%$, with the same trend on the self-collected tasks. This trade-off matches our goal of prioritizing manipulation-relevant hand-object geometry over hand pose accuracy alone.

Fig.~\ref{fig:reconstruction_comparison} further illustrates the two stages. Multi-Start Pose Refinement corrects large initial pose errors in the first two examples while preserving the plausible initialization in the third. Local Contact Refinement then addresses interpenetration, poorly localized contact, and residual hand-object gaps. Together with the improved CD$_h$, SD, and transfer success, these results show the importance of interaction geometry for downstream transfer.

\subsection{Real-World Trajectory Replay}
\label{sec:real_world_replay}

\begin{wrapfigure}{r}{0.40\textwidth}
    \vspace{-2.3\baselineskip}
    \centering
    \includegraphics[width=\linewidth]{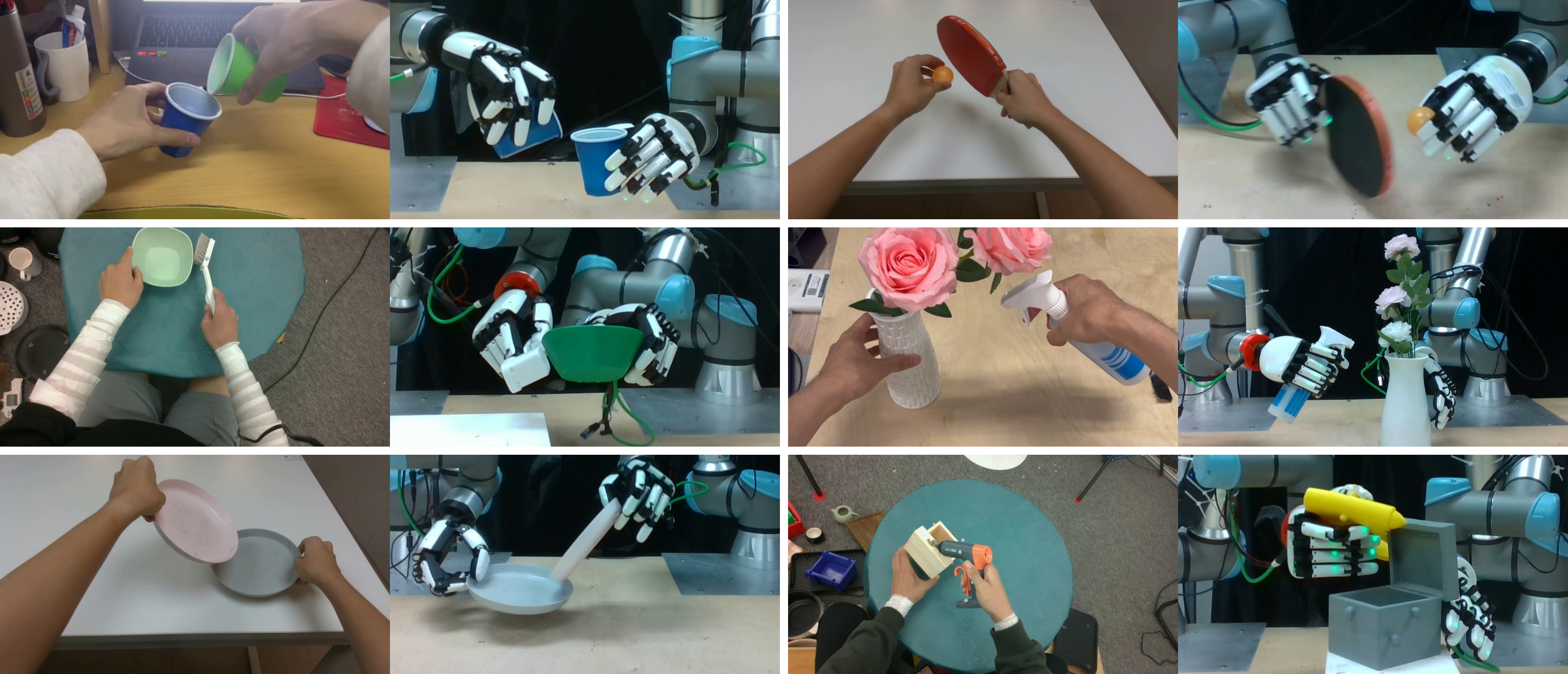}
    \captionsetup{font=small,skip=2pt}
    \caption{
    \textbf{Real-world trajectory replay.}
    Each adjacent pair shows a source human demonstration frame and the corresponding real-robot replay for the same task.
    }
    \label{fig:real_world_replay}
    \vspace{-0.3\baselineskip}
\end{wrapfigure}

We further validate the transferred trajectories on two UR5e arms equipped with XHand dexterous hands. From the trajectories generated by EgoMatrix, we select representative tasks for hardware replay, with source demonstrations drawn from both self-collected videos and public egocentric datasets. As shown in Fig.~\ref{fig:real_world_replay}, the transferred trajectories complete the corresponding manipulations on the physical system. Additional hardware details and full trajectory sequences are provided in Appendix~\ref{app:real_world_replay}. More video demonstrations are available on our project page.
\section{Conclusion and Limitations}

We presented \textbf{EgoMatrix}, an end-to-end data engine that converts monocular egocentric human videos into bimanual dexterous robot demonstrations. Contact-aware reconstruction improves the hand-object references used for manipulation transfer, while joint rendering of simulated robot and object trajectories produces physically faithful visual observations and supports data generation across task-compatible backgrounds. Experiments demonstrate improved end-to-end manipulation transfer, strong reconstruction performance, and direct trajectory replay on real robots.

\paragraph{Limitations.}
EgoMatrix's multi-stage pipeline introduces computational overhead for each demonstration. Although demonstrations can be processed in parallel with sufficient resources, further improvements in processing efficiency would facilitate larger-scale data generation. Moreover, monocular observations inherently involve depth and occlusion ambiguities. While our contact-aware reconstruction helps address these challenges, incorporating additional information, such as ground-truth depth for selected frames or multi-view observations, could further improve reconstruction quality and manipulation transfer. Finally, although we evaluate the visual fidelity of synthesized robot observations, their effectiveness for downstream manipulation policy training remains to be explored.

\bibliography{iclr2027_conference}
\bibliographystyle{iclr2027_conference}

\appendix
\section*{Appendix}
\section{Additional Details of Contact-Aware Hand-Object Reconstruction}
\label{app:reconstruction_details}

\subsection{Hand-anchored Initialization}
\label{app:hand_anchored_init}

\paragraph{Scale registration and camera consistency.}
For hand $h$ at frame $t$, let $\mathcal{P}_{h,t}$ contain pixels supported by both the observed visible-hand mask and the rendered MANO surface. The sequence-level scale in Eq.~\ref{eq:scale_alignment} is estimated by robustly aggregating
\begin{equation}
r_{h,t}(u)
=
\log z^{H}_{h,t}(u)
-
\log D^{M}_{t}(u),
\qquad
u\in\mathcal{P}_{h,t},
\end{equation}
over valid observations from both hands and all frames. The resulting factor $s_{\mathrm{seq}}$ is applied to the MoGe-3 depth and point map.

For each object, we back-project the aligned depth within its visible masks to 3D and robustly estimate its observed spatial extent over multiple frames. A single uniform scale $s_j$ is then applied to the canonical mesh $\mathcal{V}_j$ for the entire sequence. The same aligned depth is also provided as the metric-depth input to the DROID-SLAM module used by HaWoR, so that camera translation is recovered under the same scale as the reconstructed hands and objects. Object tracking is subsequently performed using the scaled meshes and aligned scene geometry.

\paragraph{Object pose initialization and tracking.}
FoundationPose \citep{wen2024foundationpose} initializes each object in its first valid frame and refines subsequent pose hypotheses. For later frames, we combine a camera-motion propagation hypothesis with rigid-motion hypotheses fitted from SpatialTracker \citep{xiao2024spatialtracker} point tracks. FoundationPose refines and scores these hypotheses to select the frame-wise object pose. Under severe hand occlusion, we avoid unconstrained full-pose recovery: translation-only refinement is used when sufficient image evidence remains, while otherwise the propagated pose is retained. This reduces abrupt pose changes when most of the object surface is unobserved.

\paragraph{Source-view alignment.}
The lightweight alignment in Eq.~\ref{eq:init_alignment} jointly considers the two hands and two manipulated objects. For $\mathcal{L}_{\mathrm{2D}}$, each entity is rendered independently and combined using depth-aware soft visibility. The loss compares the resulting visible regions with the observed masks using semantic, silhouette, boundary, and per-entity coverage terms. Ambiguous mask overlaps and invalid hand observations are excluded. When an object projection loses most of the support from its observed mask, an additional projection-capture term discourages the optimizer from moving the object out of view.

\paragraph{Pairwise relative-depth constraints.}
We partition reliable visible regions into local patches and compare the relative depths of patch pairs. Let $Z_{a,p}$ and $Z_{b,q}$ denote their current rendered depths and let $\bar{D}_{a,p}$ and $\bar{D}_{b,q}$ denote robust estimates from the aligned MoGe-3 depth. We define
\begin{equation}
\Delta Z_{ap,bq}=Z_{b,q}-Z_{a,p},
\qquad
\Delta D_{ap,bq}=\bar{D}_{b,q}-\bar{D}_{a,p}.
\end{equation}
The ordering term penalizes disagreement with the observed front--back relation,
\begin{equation}
\ell_{\mathrm{ord}}^{ap,bq}
=
w_{\mathrm{ord}}^{ap,bq}
\rho
\left(
\frac{
\left[
m_{\mathrm{ord}}
-
\operatorname{sign}(\Delta D_{ap,bq})\Delta Z_{ap,bq}
\right]_+
}
{\sigma_{\mathrm{ord}}}
\right),
\end{equation}
while the gap term penalizes disagreement in relative depth,
\begin{equation}
\ell_{\mathrm{gap}}^{ap,bq}
=
w_{\mathrm{gap}}^{ap,bq}
\rho
\left(
\frac{
\Delta Z_{ap,bq}-\Delta D_{ap,bq}
}
{\sigma_{\mathrm{gap}}}
\right).
\end{equation}
The two losses are aggregated over valid hand-hand, hand-object, and object-object patch pairs. Missing rendered support receives a fixed penalty rather than being ignored, preventing the optimizer from reducing the loss by moving an entity outside the observed region.

\paragraph{Contact-frame estimation.}
For each hand, we combine HACO predictions with the proximity of the observed hand and its manipulated object. We first project MANO vertices into the source image and identify vertices near both the hand and object masks, excluding vertices more strongly associated with other objects. Let $H_{h,t}\in[0,1]$ denote the aggregated HACO probability of these vertices, and let $P_{h,t}\in[0,1]$ denote the normalized proximity of the two masks. We compute
\begin{equation}
s_{h,t}
=
\left(
0.65H_{h,t}
+
0.35P_{h,t}
\right)P_{h,t}.
\end{equation}
We apply a five-frame median filter to the resulting scores and identify contact intervals using a support threshold of $0.38$ and a high-confidence threshold of $0.68$. Each retained interval must contain at least three high-confidence frames. Short gaps of up to eight frames are merged when the hand and object remain sufficiently close. The resulting intervals define the contact-frame set $\mathcal{T}_h^{\mathrm{c}}$ in Eq.~\ref{eq:contact_frames}.

\paragraph{Temporal regularization and support grounding.}
Frame-wise translation corrections are regularized temporally in the common world frame, while a weak initialization prior discourages unnecessary displacement. We estimate the world-up direction from dominant background surface normals. For each manipulated object, task frames outside the corresponding $\mathcal{T}_h^{\mathrm c}$ provide evidence for estimating its stable support configuration and the shared support height. The resulting object poses are adjusted to the inferred support surface during non-contact intervals, while allowing the objects to leave the surface during contact. The hand trajectories remain unchanged during grounding.

\subsection{Multi-Start Pose Refinement}
\label{app:multi_start_refinement}

\paragraph{Starting-pose selection.}
We generate candidates independently for each hand-object pair. All starts share an object baseline obtained by a short pose refinement under image, relative-depth, and support constraints. For frames $t\in\mathcal{T}_h^{\mathrm c}$, we select MANO vertices with HACO probabilities above $0.8$ and scan hand translations over $[-85,85]$\,mm at $5$\,mm intervals. We retain five starts per contact frame, including the unshifted pose. For other task frames, only the unshifted pose is used.

The MANO patches remain fixed throughout each scan. For candidate $k$ and patch $p$, we compute a histogram $\mathbf{h}_{k,p}$ of the selected intersection ranks along the camera rays, normalized by the number of vertices with valid intersections. We also record the fraction $f_{k,p}$ of rays that hit the object. The histogram uses nine bins: ranks $0,\ldots,7$ and a final bin for ranks $\geq8$; patches without hits have zero histograms. These quantities form the comparison vector $\mathbf{s}_k=\operatorname{concat}_p[\mathbf{h}_{k,p},f_{k,p}/4]$.

Let $\mathcal{S}$ contain the indices of retained starts, initially only the zero-offset candidate. While candidates remain sufficiently different, we select
\begin{equation}
k^\star
=
\arg\max_{k\notin\mathcal{S}}
\left[
\min_{\ell\in\mathcal{S}}
\frac{\|\mathbf{s}_k-\mathbf{s}_{\ell}\|_1}{\dim(\mathbf{s}_k)}
-
\lambda_q\widetilde{q}_k
\right],
\qquad
\mathcal{S}\leftarrow\mathcal{S}\cup\{k^\star\},
\label{eq:multistart_selection}
\end{equation}
where $\lambda_q=0.05$. The quality cost is $q_k=f_k\bar{g}_k+(1-f_k)c_{\mathrm{miss}}$, where $f_k$ is the overall ray-hit fraction, $\bar{g}_k$ is the mean distance between valid vertex-intersection pairs, and $c_{\mathrm{miss}}$ is a fixed cost for missed rays. We obtain $\widetilde{q}_k$ by min--max normalization over the entire scan, setting it to zero when all costs are equal. When the remaining candidates have negligible diversity, we instead select by increasing $q_k$, preferring offsets at least $10$\,mm from all retained offsets. If none satisfies this spacing, all remaining offsets are considered.

\paragraph{Candidate optimization.}
Each candidate jointly optimizes the wrist and its bound object, while the two hand-object pairs are solved independently. Translation corrections are defined in the camera frame, and rotation corrections left-multiply the initial orientation. MANO articulation and shape, object geometry and scale, and camera parameters remain fixed. Translation and rotation corrections are bounded by $100$\,mm and $10^\circ$ for the wrist, and $50$\,mm and $12^\circ$ for the object.

The source-view term in $\mathcal{L}_{\mathrm{base}}$ combines silhouette overlap and boundary alignment, using the other entity masks to exclude occluded regions. Relative-depth constraints follow Appendix~\ref{app:hand_anchored_init}, but compare only the current hand-object pair using $\widetilde{D}_t$. Bidirectional penetration penalties limit interpenetration; table constraints preserve object support and discourage hand-table penetration.

We use the contact transition in Eq.~\ref{eq:coarse_objective}. Ray targets remain fixed in object coordinates and follow the current object pose; vertices without valid ray targets use nearest-surface distance. Normal compatibility is introduced as the surface contact term replaces the ray term. Temporary pose anchors are reduced with the ray term. For task frames outside $\mathcal{T}_h^{\mathrm c}$, contact attraction is disabled, while the applicable image, depth, collision, and support constraints remain active. Temporal regularization is disabled during independent candidate optimization.

\paragraph{Candidate validation and temporal selection.}
After independent optimization, we first discard numerically invalid candidates and check contact and penetration quality. Candidates satisfying the contact-distance criterion are preferred, while penetration checks prevent strongly colliding solutions from dominating the selection. If no fully safe candidate is available, we retain the finite candidate with the smallest violation rather than discarding the frame. The zero-offset start receives no special preference, and candidates are not required to improve over the original initialization under a single global objective.

The remaining candidates are evaluated using a common geometric criterion that includes source-view consistency, relative depth, contact geometry, penetration, and support constraints. Candidate-specific ray targets and temporary initialization anchors are excluded from this score. Let $U_t^k$ denote the resulting unary cost for candidate $k$ at frame $t$. We then select a temporally coherent path using second-order dynamic programming:
\begin{equation}
\{k_t^\star\}
=
\arg\min_{\{k_t\}}
\left[
\sum_t U_t^{k_t}
+
\sum_t E_{\mathrm{pair}}(k_t,k_{t+1})
+
\sum_t E_{\mathrm{acc}}(k_{t-1},k_t,k_{t+1})
\right].
\label{eq:multistart_temporal_selection}
\end{equation}
Here, $E_{\mathrm{pair}}$ penalizes inconsistent relative hand-object states across adjacent frames, while $E_{\mathrm{acc}}$ penalizes abrupt wrist and object motion. These terms depend on the optimized geometry rather than on changes in candidate indices. The selected path is finally refined with temporal regularization in the common world frame. This refinement also covers task frames outside $\mathcal{T}_h^{\mathrm c}$, with additional motion-preserving corrections and swept collision checks where applicable.

\subsection{Local Contact Refinement}
\label{app:local_contact_refinement}

\paragraph{Stable contact construction.}
We follow the canonical-space stabilization strategy of C2Dex~\citep{ren2026c2dex}. Within $\mathcal{T}_h^{\mathrm{c}}$, temporally filtered HACO predictions are used to select candidate hand vertices, and small disconnected components are removed before ray casting. For each candidate vertex, we cast a ray from
the camera center through its initial 3D position. Among the intersections
with the bound object, we retain those within $15$\,mm in ray depth and
with opposing-normal compatibility
$-\mathbf{n}_{h}^{\top}\mathbf{n}_{o}>0.5$,
then select the intersection with the strongest normal match.

The selected points are transformed to canonical object coordinates and
grouped into locally stable interaction segments according to hand
articulation and relative hand-object motion. For each MANO vertex within
a segment, we cluster its canonical observations using DBSCAN with
$\epsilon=5$\,mm and $\mathrm{min\_samples}=3$. The dominant non-noise
cluster is retained, and its medoid defines the stable object-side target.
Using the medoid keeps the target on an observed object surface rather than
introducing an averaged point away from the mesh.

\paragraph{Opposition supplementation.}
Within $\mathcal{T}_h^{\mathrm{c}}$, when stable contacts provide insufficient opposing support, we supplement them according to the pad, palm, and side opposition types of the GRASP taxonomy~\citep{feix2016grasp}. We use a simplified semantic
partition of MANO adapted from the hand-part regions of ContactPrompt
\citep{jung2026contactprompt}. Depending on the opposition type,
supplementary vertices are drawn from the corresponding thumb, finger-pad,
palm, or side regions and prioritized by HACO confidence. Their object-side
targets are obtained from the nearest compatible surface point subject to
distance and normal constraints.

\paragraph{Optimization.}
The object trajectory and MANO shape remain fixed. For each hand, we optimize the wrist translation, wrist rotation, and all $15$ MANO joint rotations over all task frames $\mathcal{T}_h$, initialized from the Multi-Start Pose Refinement result. In implementation, the contact term in
Eq.~\ref{eq:fine_contact} is aggregated over contact patches:

\begin{equation}
\mathcal{L}_{\mathrm{contact}}
=
\frac{1}{|\mathcal{T}_h^{\mathrm{c}}|}
\sum_{t\in\mathcal{T}_h^{\mathrm{c}}}
\sum_{p}
W_{t,p}
\frac{
\sum_{i\in\mathcal{P}_{t,p}}
a_{t,i}
\left\|
\mathbf{v}_{t,i}-\mathbf{y}^{*}_{t,i}
\right\|_{2}^{2}
}{
\sum_{i\in\mathcal{P}_{t,p}} a_{t,i}
}.
\end{equation}

where $\mathcal{P}_{t,p}$ is a patch of vertices with valid contact targets, including supplementary opposition contacts when applicable. The vertex weight
$a_{t,i}$ combines HACO confidence and temporal support, while
$W_{t,p}$ controls the contribution of each patch. This prevents densely
sampled contact regions from dominating the refinement.

We penalize penetration in both directions. For hand-to-object
penetration, let $d_{o,t}(\mathbf{v})$ denote signed distance to the object
surface, positive outside:
\begin{equation}
\mathcal{L}_{h\rightarrow o}
=
\frac{1}{|\mathcal{T}_h|}
\sum_{t\in\mathcal{T}_h}
\frac{1}{N_h}
\sum_i
\left[
-d_{o,t}(\mathbf{v}_{t,i})
\right]_{+}.
\end{equation}
The reverse term penalizes sampled object-surface points that lie inside
the closed hand mesh. Additional regularization limits deviation from the
coarse wrist and articulation parameters, penalizes second-order temporal
variation, enforces anatomical joint-angle plausibility, and prevents
hand-table penetration. For frames outside $\mathcal{T}_h^{\mathrm{c}}$, contact attraction is disabled, but wrist pose and articulation are still refined using the remaining penetration and regularization terms.

We optimize each hand independently with Adam for $1000$ iterations.
Wrist translation, wrist rotation, and articulation use learning rates
$0.001$, $0.005$, and $0.0055$, with residual bounds of $60$\,mm,
$0.35$\,rad, and $0.45$\,rad per MANO joint, respectively.

\section{Details of Demonstration Synthesis and Background Transfer}
\label{app:demo_synthesis}

\subsection{Demonstration Synthesis}
\label{app:demo_synthesis}

\paragraph{Background removal and shadow transfer.}
The removal masks combine the reconstructed hand and object masks with additional arm and sleeve segmentations. We apply video inpainting with SVOR \citep{hu2026svor} and use its native prediction as the background. The simulation foreground is the union of the two dexterous-hand and manipulated-object segmentation masks.

To preserve interaction shadows, we render the scene with and without the simulated hands and objects under identical camera and lighting conditions. Their linear-image intensity ratio defines a table-region shadow transmittance $\tau_t$. We apply the resulting attenuation to the inpainted background before the binary compositing in Eq.~\ref{eq:demo_compositing}. The foreground itself is not geometrically warped or modified by a generative model.

\paragraph{Temporal correspondence.}
The synthesized video is rendered at $30$ Hz, whereas policy control and physics simulation run at $60$ Hz and $120$ Hz, respectively. Each video frame stores its associated simulator-state and source-camera indices, while the complete control-rate trajectory is retained separately. The recorded trajectory contains robot and object states, policy actions, actuator targets, wrist wrenches, and transition outcomes.

\subsection{Background Transfer}
\label{app:background_transfer}

\paragraph{Scene placement.}
Let $A$ and $B$ denote the source and target task coordinate systems. For each scene, we define a table-aligned task frame $\mathbf{F}$ whose origin is the midpoint of the two initial object positions projected onto the support plane. Its $x$-axis follows the direction between the two objects, while the $z$-axis is the world-up direction. The initial source-to-target placement is
\begin{equation}
\mathbf{T}^{0}_{B\leftarrow A}
=
\mathbf{F}_{B}\mathbf{F}_{A}^{-1}.
\end{equation}
An optional fixed in-plane translation $\mathbf{D}_{A}$ may be applied for framing,
\begin{equation}
\mathbf{T}_{B\leftarrow A}
=
\mathbf{T}^{0}_{B\leftarrow A}\mathbf{D}_{A}.
\label{eq:background_scene_transform}
\end{equation}
The same transform is shared by both hands and all manipulated objects. No scale change or frame-wise alignment is applied.

Rather than rewriting the recorded source states, we equivalently express the target camera in the source task frame. If $\mathbf{C}_{B}(j)$ is the target camera-to-world pose at target frame $j$, the rendering camera for output frame $i$ is
\begin{equation}
\mathbf{C}_{A}(i)
=
\mathbf{T}_{B\leftarrow A}^{-1}
\mathbf{C}_{B}\!\left(j(i)\right).
\label{eq:background_target_camera}
\end{equation}
Target intrinsics are used for rendering, and the target support proxy is transformed consistently with the camera and scene placement.

\paragraph{Temporal mapping.}
The source execution remains on its original $60$ Hz state and control trajectory, while the synthesized video is rendered at $30$ Hz. We therefore use every second recorded source state. When the available target-background segment is shorter than the source output, target RGB frames and camera poses are sampled uniformly over the available segment using the same frame mapping; the source execution itself is not time-warped.

The target support representation is taken from the configured background scene rather than re-estimated during background transfer. Accordingly, this stage preserves the original executed hand-object geometry and its motion relative to a compatible support plane, but does not re-simulate contact dynamics in the target scene.

\section{Reconstruction Evaluation Protocol}
\label{app:reconstruction_evaluation}

We report ten reconstruction metrics: MPJPE$_{\mathrm{RA}}$, CD, F@5,
ATE, ARE, RTE, RRE, CD$_h$, SD, and
$D_{\mathrm{pen}}^{\max}$. Unless otherwise specified below, all methods
are evaluated using the same metric implementation and aggregation
protocol.

\paragraph{Hand pose and object geometry.}
Following \citet{fan2024hold}, hand pose is evaluated using
MPJPE$_{\mathrm{RA}}$ over $21$ MANO keypoints. Prediction and ground
truth are independently translated to their wrist roots before comparison;
no rotational, scale, or Procrustes alignment is applied.

For object geometry, we uniformly sample $20{,}000$ surface points by
triangle area from both the reconstructed and ground-truth canonical
meshes. A proper $\mathrm{Sim}(3)$ transformation, consisting of rotation,
translation, and uniform scale without reflection, is estimated once per
object and kept fixed. CD is the sum of the two mean nearest-surface
Euclidean distances and is therefore reported in centimeters rather than
squared centimeters. F@5 uses a $5$\,mm nearest-surface threshold.

\paragraph{Object trajectory.}
We evaluate object motion using ATE, ARE, RTE, and RRE following
\citet{shin2026compose}. To avoid dependence on arbitrary canonical object
origins, translation trajectories are represented by the area-weighted
surface centroids of the posed meshes.

For each demonstration, all manipulated objects share a single
fixed-scale rigid alignment between the predicted and ground-truth worlds.
The alignment is estimated jointly from all valid object-center
correspondences, with each object given equal total weight. No scale
adjustment is allowed. This alignment is estimated once per demonstration
and is never re-estimated independently for individual objects or frames.

The canonical orientation of a reconstructed object can differ from that
of its ground-truth mesh by a constant rotation. We therefore remove one
fixed orientation offset using the first common valid frame of each object
and keep it unchanged throughout the sequence. No frame-wise rotational
alignment is applied. ARE and RRE use the standard geodesic distance on
$SO(3)$; object symmetry is not explicitly modeled. RTE and RRE are
computed only between consecutive valid prediction frames and never bridge
missing frames.

DexImit produces trajectories at a temporal stride of three frames.
Its trajectory metrics are therefore evaluated on its native sampled
frames, and its relative errors correspond to three-frame intervals; these
entries are marked accordingly in the table.

MagicHOI reconstructs its hand-object branches under independent SfM
gauges and does not provide a common reconstructed camera trajectory.
For MagicHOI only, we use the ground-truth camera trajectory to transform
its valid camera-frame object predictions into the ground-truth world
before computing the same trajectory metrics. These GT-camera-assisted
results are marked with $^\ast$ in the table.

\paragraph{Hand-object interaction.}
We evaluate relative hand-object geometry using CD$_h$
\citep{fan2024hold}. At each valid frame, the predicted and ground-truth
posed object surfaces are expressed relative to their corresponding wrist
roots in the camera frame, after which the same unsquared bidirectional
Chamfer distance as above is computed. No additional per-frame scale,
rotation, or rigid alignment is applied.

Physical stability is evaluated using a GraspTTA-style simulation
displacement (SD) protocol \citep{jiang2021grasptta}. For each evaluated
hand-object configuration, the hand is held static and the object is
simulated dynamically in PyBullet for $35$ steps at $1/240$\,s. The same
gravity, object mass, contact, solver, and collision settings are used for
all methods. SD measures the displacement of the simulated object between
its initial and final states.

We additionally report maximum penetration depth
$D_{\mathrm{pen}}^{\max}$ \citep{hasson2019joint}. For each frame, we
measure the largest penetration of MANO vertices into the posed object
surface; frames without penetration contribute zero. The reported value is
the mean of these frame-wise maxima over the evaluated sequence.

\paragraph{Aggregation and validity.}
Metrics are aggregated with equal weighting across the relevant hands or
objects within each demonstration, followed by equal weighting across
demonstrations. Geometry metrics are computed once for each reconstructed
canonical object, whereas pose and interaction metrics are evaluated over
their valid temporal support. Missing predictions are not interpolated.
When a method does not produce the output required by a metric, the
corresponding entry is reported as N/A. For methods with incomplete
reconstruction coverage, we additionally report the number of valid
demonstrations or branches used for evaluation.


\section{Visual Synthesis Baseline Protocols}
\label{app:visual_baselines}

\paragraph{Source-background synthesis.}
For Kiwi-Edit~\citep{lin2026kiwiedit}, we provide the complete source human-manipulation video together with a reference image showing XHand dexterous hands mounted on Rainbow Robotics RB-Y1 arms. We use the following instruction for all videos:
\begin{quote}
\small
``Replace the human hands and arms with XHand dexterous hands mounted on Rainbow Robotics RB-Y1 robot arms, performing the same manipulation task. Keep the original background and all other scene elements unchanged.''
\end{quote}
The generated video is used directly without additional geometric correction or post-processing.

For GPT-6 Astra~\citep{openai2026gpt6astra}, each source frame is edited using the human frame as the manipulation reference and the same RB-Y1--XHand reference image(s) used to specify the target embodiment. We use the following fixed prompt:
\begin{quote}
\small
``Edit the first image. Replace the visible human hand(s) manipulating the object with XHand dexterous hand(s) mounted on a Rainbow Robotics RB-Y1 dual-arm robot, matching the reference image(s). If any wrist, forearm, or arm is visible, replace it with the corresponding RB-Y1 robot arm segment and keep the XHand-arm connection physically plausible. Preserve plausible hand-object contact. Do not add text, extra objects, or a collage layout.''
\end{quote}
We retain the first generated result for each frame and perform no manual correction or result selection.

\paragraph{Target-background synthesis.}
For the target-background experiment, GPT-6 Astra receives three visual inputs for each sampled frame: the human source frame, one fixed image of the target real-robot workspace, and one fixed robot reference image based on the same UR5e and XHand assets used in EgoMatrix. The source frame specifies the manipulation action, manipulated objects, hand-object interaction, and interaction stage, while the fixed target image specifies the camera viewpoint, workspace, background, and overall composition. The robot and object states already present in the target image are not treated as references for the desired manipulation. The robot reference specifies the robot geometry, proportions, mounting, and colors, rather than the desired pose or scene layout.

The target embodiment is fixed to two UR5e arms, each equipped with an XHand dexterous hand. We use the same target image, robot reference image, and following prompt for every source frame, with the three images supplied in the order described above:
\begin{quote}
\small
``Use image 1, the current source frame, as the reference for the manipulation action, manipulated objects, hand-object interaction, and interaction stage. Edit image 2, the fixed target workstation image, so that a bimanual robot consisting of two XHand dexterous hands, each mounted on a UR5e robot arm, performs the same manipulation shown in image 1. Image 3 is a fixed visualization of the supplied assembled UR5e + XHand assets; use it only as a reference for the robot geometry, proportions, mounting, and colors, not for the robot pose, camera, lighting, background, or manipulation state.

Match the manipulated objects, their poses, the grasp or contact configurations, and the stage of the manipulation to image 1 as closely as possible, expressed from image 2's camera viewpoint. Replace the robot configuration and manipulated-object state already visible in image 2 whenever necessary; they do not define the desired action. Reproduce both hands' involvement when the source manipulation is bimanual. If only one hand is visibly involved, do not force the second hand to become prominent, but keep the embodiment consistent with two UR5e arms and two mounted XHands.

Preserve image 2's camera viewpoint, workspace, background, table, overall scene composition, and visual appearance. Match the robot embodiment shown in image 3 while depicting it under the target workstation's illumination. The UR5e arm segments must connect naturally to the XHand hands and remain physically plausible from the target viewpoint.

Do not copy image 1's viewpoint or background. Do not preserve image 2's robot pose when it conflicts with the source manipulation. Do not copy image 3's demonstration pose or backdrop. Do not add text, unrelated objects, extra robot limbs, or a collage layout. Return a single edited target-workstation image at 1920x1080 pixels in landscape 16:9 format.''
\end{quote}

Each frame is generated independently with the same prompt and fixed reference images. We use one generation per frame, retain the first result, and apply no manual selection or post-processing. No EgoMatrix output, previously generated Astra frame, or additional reference image is provided.

\section{Robot Deployment}
\label{app:real_world_replay}

\begin{figure}[H]
    \centering

    \begin{subfigure}[t]{0.485\linewidth}
        \centering
        \includegraphics[width=\linewidth]{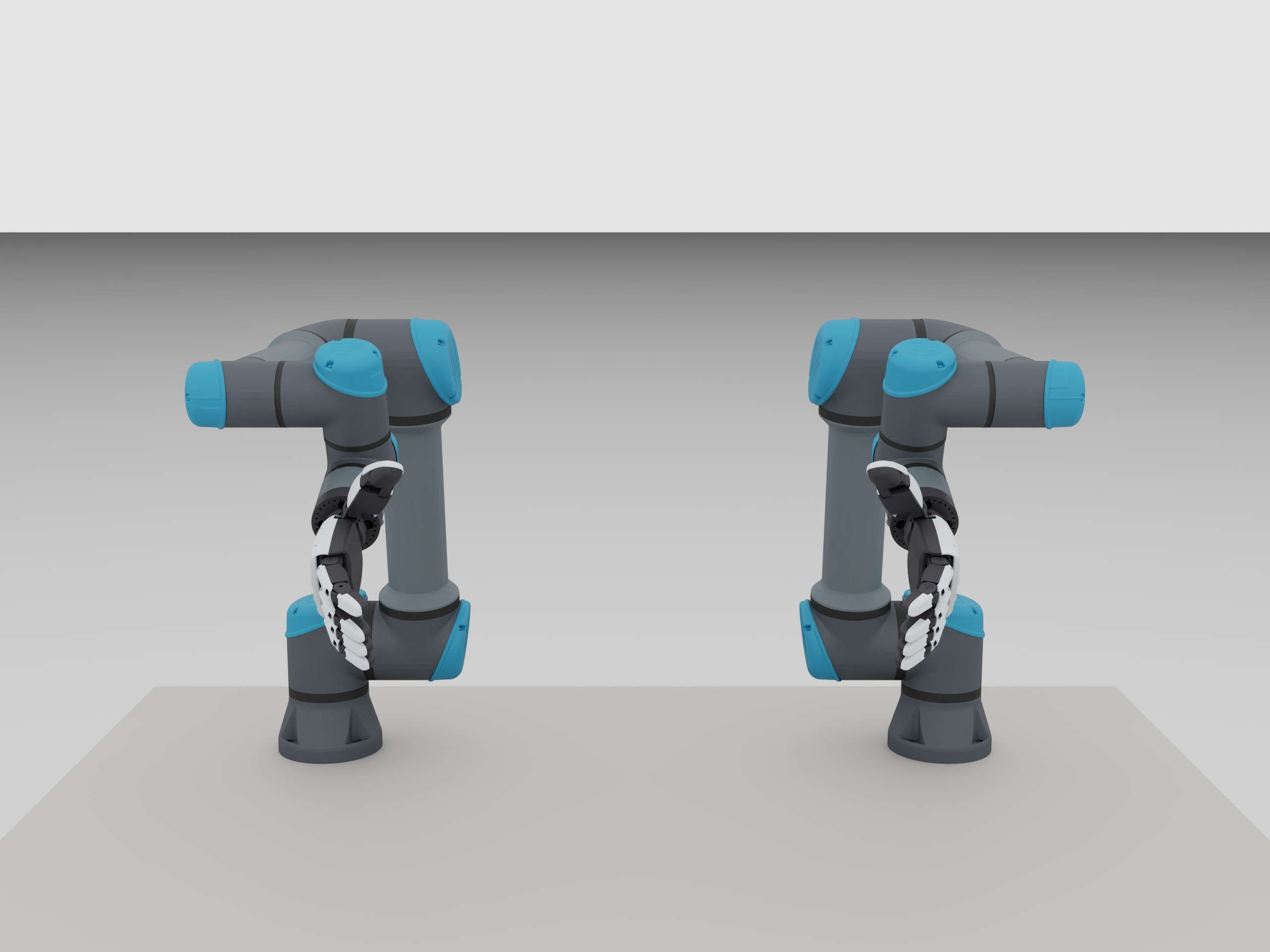}
        \caption{Dual UR5e arms with XHand dexterous hands.}
        \label{fig:robot_platform_ur5e}
    \end{subfigure}
    \hfill
    \begin{subfigure}[t]{0.485\linewidth}
        \centering
        \includegraphics[width=\linewidth]{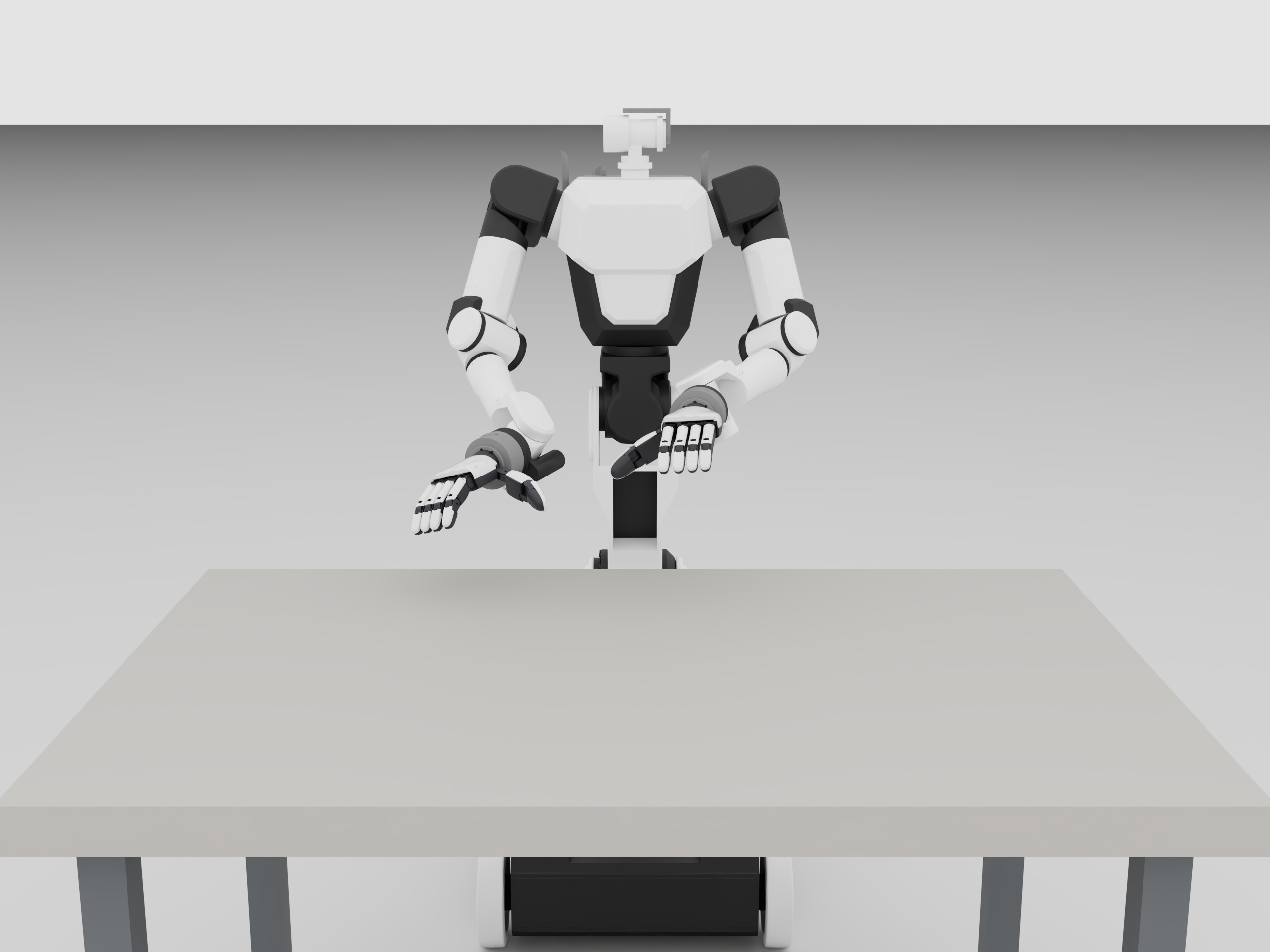}
        \caption{Rainbow Robotics RB-Y1 with XHand.}
        \label{fig:robot_platform_rby1}
    \end{subfigure}

    \captionsetup{font=small}
    \caption{
        \textbf{Real-robot platforms used for trajectory replay.}
        We evaluate transferred trajectories on two embodiments equipped with XHand dexterous hands: a dual-UR5e setup and the dual-arm Rainbow Robotics RB-Y1 platform.
    }
    \label{fig:robot_platforms}
\end{figure}

We deploy the transferred trajectories on two real-robot platforms. The first platform consists of two UR5e arms, each equipped with an XHand dexterous hand. The second uses the dual-arm system of a Rainbow Robotics RB-Y1, also equipped with XHand dexterous hands. Fig.~\ref{fig:robot_platforms} shows the two hardware setups used in our experiments.

\begin{figure}[t]
    \centering
    \includegraphics[width=\linewidth]{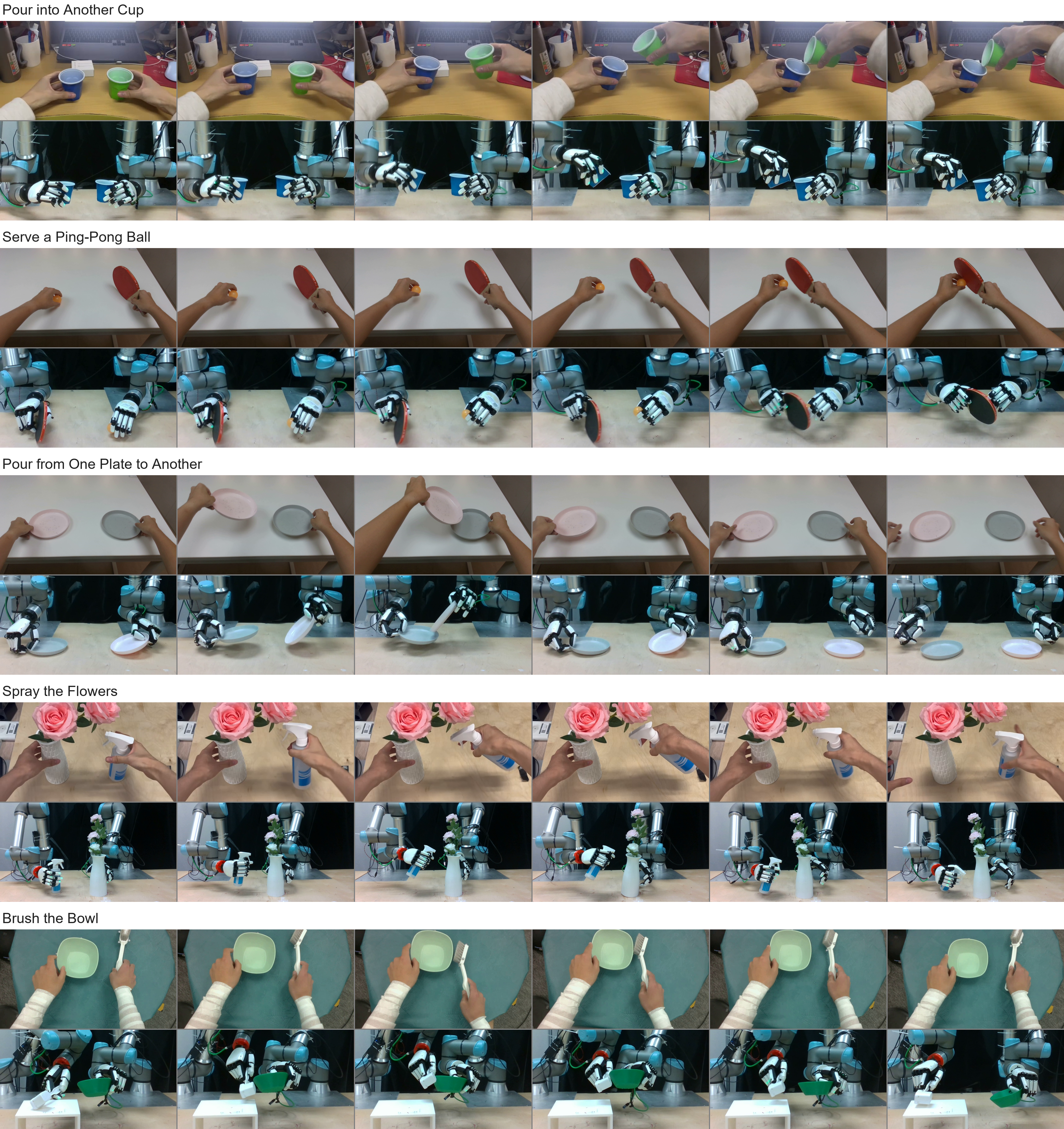}

    \captionsetup{font=small}
    \caption{
        \textbf{Real-world trajectory replay across representative tasks.}
        For each task, the top row shows the source human demonstration and the bottom row shows the corresponding real-robot replay. Time progresses from left to right.
    }
    \label{fig:real_world_full_sequences}
\end{figure}

Fig.~\ref{fig:real_world_full_sequences} shows five representative real-world replay sequences. For each task, the upper row contains frames sampled from the source human demonstration, while the lower row shows the corresponding stages of the transferred trajectory replayed on the real robot. The examples include demonstrations collected by us as well as demonstrations from public egocentric datasets.

\end{document}